\documentclass[9pt,twoside]{extarticle}
\usepackage[papersize={17cm,24cm},margin=22mm,top=17mm,headsep=5mm]{geometry} 
\usepackage{amssymb}
\usepackage{amsmath}
\usepackage{amsthm}
\usepackage{enumerate}
\usepackage{verbatim} % verbatim input
\usepackage{titling} % customize title
\usepackage{csquotes} % context sensitive quotes
\usepackage{upquote} % allow correct quotes in verbatim
\usepackage[bottom]{footmisc} % put footnotes down at the bottom margin
\usepackage{graphicx} % \includegraphics
\usepackage[noend]{algpseudocode}
\usepackage[linesnumbered,ruled]{algorithm2e}
\usepackage{thmtools}
\usepackage{thm-restate}
\usepackage[dvipsnames]{xcolor}
\usepackage[colorlinks,linkcolor = Aquamarine, urlcolor  = violet, citecolor = YellowOrange, anchorcolor = violet]{hyperref}

\usepackage[english]{babel} % multilinguality
\date{}
\usepackage{natbib}
\renewenvironment{abstract}
  {\noindent\small %\quotation
  {\noindent{\large\textbf\abstractname. }%\par\nobreak\smallskip
  \thispagestyle{plain}
  }}
  {}
    
\usepackage{fancyhdr}
\fancypagestyle{plain}{% first page of chapter
  \fancyhead[C,R]{}
  \fancyfoot{} % comment to put page number at bottom
  } % load packages, set parameters, define commands
\newcommand{\BlackBox}{\rule{1.5ex}{1.5ex}}  % end of proof
\newenvironment{Proof}{\par\noindent{\bf Proof\ }}{\hfill\BlackBox\\[2mm]}
\newtheorem{theorem}{Theorem}

\newtheorem{lemma}[theorem]{Lemma}

\newtheorem{remark}[theorem]{Remark}
\newtheorem{corollary}[theorem]{Corollary}

\def\ci{\perp\!\!\!\perp}

\def\lv{\lVert}
\def\rv{\rVert}

\title{Alternating minimization for dictionary learning: Local Convergence Guarantees}
\date{\today}
\author{
Niladri S. Chatterji\\
University of California, Berkeley\\
\texttt{chatterji@berkeley.edu}\\
\and
Peter L. Bartlett\\
University of California, Berkeley\\
\texttt{peter@berkeley.edu}\\
}

\begin{document}
\maketitle

\begin{abstract}
We present theoretical guarantees for an alternating minimization
algorithm for the dictionary learning/sparse coding problem. The
dictionary learning problem is to factorize vector samples
$y^{1},y^{2},\ldots, y^{n}$ into an appropriate basis (dictionary)
$A^*$ and sparse vectors $x^{1*},\ldots,x^{n*}$. Our algorithm is a
simple alternating minimization procedure that switches between
$\ell_1$ minimization and gradient descent in alternate steps.
Dictionary learning and specifically alternating minimization
algorithms for dictionary learning are well studied both theoretically
and empirically. However, in contrast to previous theoretical analyses
for this problem, we replace a condition on the operator norm
(that is, the largest magnitude singular value) of the true underlying
dictionary $A^*$ with a condition on the matrix infinity norm (that
is, the largest magnitude term). Our guarantees are under a reasonable
generative model that allows for dictionaries with growing operator
norms, and can handle an arbitrary level of overcompleteness, while having
sparsity that is information theoretically optimal. We also establish
upper bounds on the sample complexity of our algorithm.
\end{abstract}

\paragraph{Erratum.} \emph{An earlier version of this paper appeared in NIPS 2017 which had an erroneous claim about convergence guarantees with random initialization. The main result -- Theorem \ref{maintheorem} -- has been corrected by adding an assumption about the initialization (Assumption B1).}
\section{Introduction}
In the problem of sparse coding/dictionary learning, given i.i.d. samples $y^{1},y^2,\ldots,y^{n} \in \mathbb{R}^{d}$ produced from the generative model 
\begin{align}
y^{i} & = A^{*}x^{i*} 
\end{align} 
for $i \in \{1,2,\ldots, n\}$, the goal is to recover a fixed
dictionary $A^{*} \in \mathbb{R}^{d \times r}$ and $s$-sparse vectors
$x^{i*} \in \mathbb{R}^{r}$. (An $s$-sparse vector has no more than
$s$ non-zero entries.) In many problems of interest, the
dictionary is often overcomplete, that is, $r\ge d$. This is believed to
add flexibility in modeling and robustness. This model was first
proposed in neuroscience as an energy minimization
heuristic that reproduces features of the V1 region of the visual cortex
\citep{ols, lew}. It has also been an extremely
successful approach to identifying low dimensional structure in high
dimensional data; it is used extensively to find features in images,
speech and video \citep[see, for example, references in][]{elad}. 

Most formulations of dictionary learning tend to yield non-convex
optimization problems. For example, note that if either $x^{i*}$ or
$A^*$ were known, given $y^{i}$, this would just be a (matrix/sparse)
regression problem. However, estimating both $x^{i*}$ and $A^*$
simultaneously leads to both computational as well as statistical
complications. The heuristic of
alternating minimization works well empirically for dictionary
learning. At each step, first an
estimate of the dictionary is held fixed while the sparse coefficients
are estimated; next, using these sparse coefficients the dictionary is
updated. Note that in \emph{each step} the sub-problem has a convex
formulation, and there is a range of efficient algorithms that can be
used. This heuristic has been very successful empirically, and there
has also been significant recent theoretical progress in understanding
its performance, which we discuss next.

\subsection{Related Work}
A recent line of work theoretically analyzes local linear convergence
rates for alternating minimization procedures applied to dictionary
learning \citep{agarwal1,arora1}. \citet{arora1} present a neurally
plausible algorithm that recovers the dictionary exactly for sparsity
up to $s = \mathcal{O}(\sqrt{d}/(\mu \log(d)))$, where $\mu/\sqrt{d}$ is the
level of incoherence in the dictionary (which is a measure of the
similarity of the columns; see Assumption~A\ref{assumption:incoherence}
below).
\citet{agarwal1} analyze a least squares/$\ell_1$ minimization scheme
and show that it can tolerate sparsity up to
$s = \mathcal{O}(d^{1/6})$. Both of
these establish local linear convergence guarantees for the maximum
column-wise distance. Exact recovery guarantees require a
singular-value decomposition (SVD) or clustering based procedure to
initialize their dictionary estimates \citep[see also the previous
work~][]{arora2,agarwal2}.

For the undercomplete case (when $r\le d$), \citet{sun} provide a
Riemannian trust region method that can tolerate sparsity $s =
\mathcal{O}(d)$, while earlier work by \citet{spielman} provides an
algorithm that works in this setting for sparsity
$\mathcal{O}(\sqrt{d})$.

Local and global optima of non-convex formulations for the problem
have also been extensively studied in \citep{wu, sparsespur,
gribonval}, among others. Apart from alternating minimization, other
approaches (without theoretical convergence guarantees) for dictionary
learning include K-SVD \citep{aharon} and MOD \citep{MOD}. There is
also a nice formulation by \citet{barak},
based on the sum-of-squares hierarchy. Recently, \citet{hazan} provide
guarantees for improper
dictionary learning, where instead of learning a dictionary, they learn
a comparable encoding via convex relaxations. Our work also adds to
the recent literature on analyzing alternating minimization algorithms
\citep{jain2,jain3,jain4,hardt,bal}.
%and the related analysis of the EM algorithm \citep{bal}.
\subsection{Contributions}
Our main contribution is to present new conditions under which
alternating minimization for dictionary learning converges at a linear
rate to the optimum. We impose a condition on the matrix
infinity norm (largest magnitude entry) of the underlying
dictionary. This allows dictionaries with
operator norm growing with
dimension ($d,r$). The error rates are measured in the matrix infinity
norm, which is sharper than the previous error rates in maximum
column-wise error.

Our results hold for a rather arbitrary
level of overcompleteness, $r = \mathcal{O}(poly(d))$. We establish
convergence results for sparsity level $s = \mathcal{O}(\sqrt{d})$,
which is information theoretically optimal for incoherent dictionaries
and improves the previously best known results in the overcomplete setting
by a logarithmic factor. Our algorithm is simple,
involving an $\ell_1$-minimization step followed by a gradient
update for the dictionary.

A key step in our proofs is an analysis of a robust sparse
estimator---$\{\ell_1,\ell_2,\ell_{\infty} \}$-MU Selector---under
fixed (worst case) corruption in the dictionary. We prove that this
estimator is minimax optimal in this setting,
which might be of independent interest.
%We believe this analysis of the robust sparse estimator maybe of independent interest apart from its application to dictionary learning presented here. 
\subsection{Organization}%In the next section we establish the notation that we use throughout the paper. 
In Section \ref{algosec}, we present our alternating minimization
algorithm and discuss the sparse regression estimator. In Section
\ref{mainresultssec}, we list the assumptions under which our
algorithm converges and state the main convergence result. Finally,
in Section~\ref{convergencesec}, we prove convergence of our
algorithm. We defer
technical lemmas, analysis of the sparse regression estimator, and
minimax analysis to the appendix.
\subsubsection*{Notation}
For a vector $v \in \mathbb{R}^{d}$, $v_{i}$ denotes the $i^{th}$
component of the vector, $\lv v \rv_{p}$ denotes the $\ell_{p}$ norm,
$supp(v)$ denotes the support of a vector $v$, that is, the
set of non-zero entries of the vector, $sgn(v)$ denotes the sign of the
vector $v$, that is, a vector with $sgn(v)_{i} =1$ if $v_{i} > 0$,
$sgn(v)_{i} =-1$ if $v_{i} < 0$ and  $sgn(v)_{i}=0$ if $v_{i} = 0$.
For a matrix $W$, $W_i$ denotes the $i^{th}$ column, $W_{ij}$ is
the element in the $i^{th}$ row and $j^{th}$ column, $\lv
W \rv_{op}$ denotes the operator norm, and $\lv W \rv_{\infty}$ denotes the
maximum of the magnitudes of the elements of $W$.
For a set $J$, we denote its cardinality by $\lvert J \rvert$.
Throughout the paper, we
use $C$ multiple times to denote global constants that are independent
of the problem parameters and dimension.
%We will also use $g(n) = \mathcal{O}^*(f(n))$ to signify that $g(n)$ is upper bounded by $C f(n)$ for some small enough constant $C$ and $g(n) = \Omega^*(f(n))$ to signify that $g(n)$ is lower bounded by $C f(n)$ for some constant $C$. 
We denote the indicator function by $\mathbb{I}(\cdot)$.

\section{Algorithm}
\label{algosec}\begin{algorithm}[t] 
\caption{Alternating Minimization for Dictionary Learning} 
\label{Infsample}
\SetKwInOut{Input}{Input}
    \SetKwInOut{Output}{Output}
    \Input{Step size $\eta $, samples $\{ y^k\}_{k=1}^{n}$, initial estimate $A^{(0)}$, number of steps $T$, thresholds $\{\tau^{(t)}\}_{t=1}^{T}$, initial radius $R^{(0)}$ and parameters $\{\gamma^{(t)}\}_{t=1}^T,\{\lambda^{(t)}\}_{t=1}^T \text{ and } \{\nu^{(t)}\}_{t=1}^T$.}
    
    \For {$t=1,2,\ldots,T$} 
      {
      \For {$k=1,2,\ldots, n$} 
      {
         $w^{k,(t)} = MUS_{\gamma^{(t)},\lambda^{(t)},\nu^{(t)}}(y^k, A^{(t-1)},R^{(t-1)})$
         
          \For {$l=1,2,3 \ldots, r$} 
      		{   
	      $x_{l}^{k,(t)}  = w_{l}^{k,(t)} \mathbb{I} \left( \lvert w_{l}^{k,(t)}\rvert > \tau^{(t)} \right), \quad (x^{k,(t)} \text{ is the sparse estimate})$
     		}
   	  }
   	  \For {$i=1,2,\ldots, d$}
   	  {
   	  \For {$j=1,2,\ldots, r $}
   	  {
      $A_{ij}^{(t)} = A_{ij}^{(t-1)} - \frac{\eta}{n} \sum_{k=1}^{n}\left[\sum_{p=1}^{r} \left(A_{ip}^{(t-1)} x^{k,(t)}_p x^{k,(t)}_j  - y^k_i x^{k,(t)}_j\right) \right] $
     }
     }
     $R^{(t)} = \frac{7}{8} R^{(t-1)}$.
   }
   \end{algorithm}   
Given an initial estimate of the dictionary $A^{(0)}$ we alternate
between an $\ell_1$ minimization procedure (specifically the
$\{\ell_1,\ell_2,\ell_{\infty}\}$-MU Selector---$MUS_{\gamma,\lambda,\nu}$ in
the algorithm---followed by a thresholding step) and a gradient step,
under sample $\ell_2$ loss, to update the dictionary. We analyze this algorithm and demand linear convergence at a rate of
7/8; convergence analysis for other rates follows in the same vein
with altered constants. Below we state the permitted range for the various parameters in the algorithm above.
\begin{enumerate}
\item Step size: We need to set the step size in the range $3r/4s < \eta < r/s$.
\item Threshold: At each step set the threshold at $\tau^{(t)} = 16R^{(t-1)}M(R^{(t-1)}(s+1)+s/\sqrt{d})$.
\item Tuning parameters: We need to pick $\lambda^{(t)}$ and $\nu^{(t)}$ such that the assumption (D5) is satisfied. A choice that is suitable that satisfies this assumption is
\begin{align*}
128 s \left(R^{(t-1)}\right)^2\le \nu^{(t)} &\le 3,\\
32\left(s^{3/2}\left(R^{(t-1)}\right)^2+\frac{s^{3/2}R^{(t-1)}}{d^{1/2}} \right)\left(4+ \frac{6}{\sqrt{s}} \right) \le \lambda^{(t)}  & \le 3.
\end{align*}
We need to set $\gamma^{(t)}$ as specified by Theorem \ref{bell},
\begin{align*}
\gamma^{(t)} & = \sqrt{s}\left(R^{(t-1)}\right)^2 + \sqrt{\frac{s}{d}}R^{(t-1)}.
\end{align*}
\end{enumerate}
%We note that other sparse recovery estimators could also be used instead like OMP \citep{chen}, non-convex LASSO estimator \citep{loh} and the MU-Selector \citep{MU} however these would lead to worse guarantees for the radius of convergence. 

%We show in the following section that $\{\ell_1,\ell_2,\ell_{\infty}\}$-MU Selector is minimax optimal (errors scaling with $\lv \theta^* \rv_{2}$ rather than $\lv \theta^* \rv_1$ where $\theta^*$ is the true underlying sparse vector) when we have an inaccurate estimate of the dictionary. This leads to a tight bound on the radius of convergence and improved convergence guarantees. We could also consider an algorithm where the gradient step is replaced by least squares step and analyze in a similar way as \citet{agarwal1}.

\subsection{Sparse Regression Estimator}
Our proof of convergence for Algorithm \ref{Infsample} also goes through with a different choices of robust sparse regression estimators, however, we can establish the tightest guarantees when the $\{\ell_1,\ell_2,\ell_{\infty}\}$-MU Selector is used in the sparse regression step. The $\{\ell_1,\ell_2,\ell_{\infty}\}$-MU Selector \citep{bell} was established as a modification of the Dantzig selector to handle uncertainty in the dictionary. There is a beautiful line of work that precedes this that includes \citep{MU,MUimprov,belloniconic}. There are also modified non-convex LASSO programs that have been studied \citep{loh} and Orthogonal Matching Pursuit algorithms under in-variable errors \citep{chen}. However these estimators require the error in the dictionary to be stochastic and zero mean which makes them less suitable in this setting. Also note that standard $\ell_1$ minimization estimators like the LASSO and Dantzig selector are highly unstable under errors in the dictionary and would lead to much worse guarantees in terms of radius of convergence \citep[as studied in][]{agarwal1}.  
We establish the error guarantees for a robust sparse estimator  $\{\ell_1,\ell_2,\ell_{\infty}\}$-MU Selector under fixed corruption in the dictionary. We also establish that this estimator is minimax optimal when the error in the sparse estimate is measured in infinity norm $\lv \hat{\theta} - \theta^* \rv_{\infty}$ and the dictionary is corrupted. 
\subsubsection*{The $\{\ell_1,\ell_2,\ell_{\infty}\}$-MU Selector}
Define the estimator $\hat{\theta}$ such that $(\hat{\theta},\hat{t},\hat{u}) \in \mathbb{R}^{r} \times \mathbb{R}_{+} \times \mathbb{R}_{+}$ is the solution to the convex minimization problem 
\begin{align}
\label{belloni}
\min_{\theta,t,u} \Bigg\{ \lv \theta \rv_1+ \lambda t + \nu u \ \Bigg \lvert \  
\theta \in \mathbb{R}^r, \Big\lv \frac{1}{d}A^{\top}\big(y-A\theta \big)\Big\rv_{\infty} \le \gamma t + R^2 u, \lv \theta\rv_{2} \le t, \lv \theta \rv_{\infty}\le u \Bigg\}
\end{align}
where, $\gamma, \lambda$ and $\nu$  are tuning parameters that are
chosen appropriately. $R$ is an upper bound on the error in our
dictionary measured in matrix infinity norm. Henceforth the first
coordinate ($\hat{\theta}$) of this estimator is called
$MUS_{\gamma,\lambda,\nu} (y,A,R)$, where the first argument is the
sample, the second is the matrix, and the third is the value of the upper
bound on the error of the dictionary measured in infinity norm. We
will see that under our assumptions we will be able to establish an
upper bound on the error on the estimator, $ \lv \hat{\theta} -
\theta^* \rv_{\infty} \le 16 R M\left(R(s+1) +s/\sqrt{d}\right)$,
where $\lvert \theta^*_{j} \rvert \le M  \ \forall j$. We define a
threshold at each step $\tau = 16 R M(R(s+1) +s/\sqrt{d})$. The
thresholded estimate $\tilde{\theta}$ is defined as
\begin{align}
\label{thres}
\tilde{\theta}_{i} & = \hat{\theta}_{i} \mathbb{I}[ \lvert \hat{\theta}_{i}\rvert > \tau ] \qquad \forall i \in \{1,2,\ldots, r\}.
\end{align}
Our assumptions will ensure that we have the guarantee $sgn(\tilde{\theta}) = sgn(\theta^{*})$. This will be crucial in our proof of convergence. The analysis of this estimator is presented in Appendix \ref{estimator}.

%As with previous analyses for alternating minimization for dictionary
%learning \citep{agarwal1,arora1}, we rely on identifying the sign of
%the sparse covariates correctly at each step. For the sparse recovery
%step, we restrict ourselves to a class of two step estimators, where
%we first estimate a vector $\hat{\theta}$ with some error guarantee in
%infinity norm $\lv \hat{\theta} - \theta^* \rv_{\infty} \le \tau$ and
%then have an element-wise thresholding step ($\tilde{\theta}_{i}  =
%\hat{\theta}_{i} \mathbb{I}[ \lvert \hat{\theta}_{i}\rvert > \tau
%]$). To identify the sign correctly using this class of thresholded
%estimators we would like in the first step to use an estimator that is
To identify the signs of the sparse covariates correctly using this class of thresholded estimators, we would like in the first step to use an estimator $\hat\theta$ that is \emph{optimal}, as this would lead to tighter control over the radius
of convergence. This makes the choice of
$\{\ell_1,\ell_2,\ell_{\infty}\}$-MU Selector natural, as we will show
it is minimax optimal under certain settings.
\begin{theorem}[informal] \label{minimaxtheo} 
Define the sets of matrices $\mathcal{A} = \{ B\in \mathbb{R}^{d
\times r}  \big\lvert \lv B_{i} \rv_2 \le 1, \ \forall i \in
\{1,\ldots,r \} \}$ and $\mathcal{W} = \{ P \in \mathbb{R}^{d \times
r}  \big\lvert \lv P \rv_{\infty} \le R \}$ with $R =
\mathcal{O}(1/\sqrt{s})$. Then there exists an $A^* \in \mathcal{A}$
and $W \in \mathcal{W}$ with $A
\triangleq A^* + W$ such that
\begin{align}
\inf_{\hat{T}} \sup_{\theta^*} \lv \hat{T} - \theta^* \rv_{\infty} \ge C R L \left(\sqrt{1 - \frac{\log(s)}{\log(r)}} \right),
\end{align}
where the $\inf_{\hat{T}}$ is over all measurable estimators $\hat{T}$
with input $(A^*\theta^*,A,R)$, and the $sup$ is over $s$-sparse vectors
$\theta^*$ with $2$-norm $L>0$.
\end{theorem}
%\begin{theorem}[informal]\label{minimaxtheo}
%Let $y = A^*\theta^*$ where  $A^* \in \mathbb{R}^{d \times r}$ with normalized columns $\lv A^*_{i} \rv_2 \le 1$ and $\theta^*$ is a $s$-sparse vector with norm $\lv \theta^* \rv_2$. Also let $W \in \mathbb{R}^{d \times r}$ be a perturbation to this dictionary,  with $\lv W \rv_{\infty} \le R = \mathcal{O}(1/\sqrt{s})$. Finally define $A \triangleq A^* + W$, then there exists such an $A^*$ and $W$ such that for all (measurable) estimators $\hat{T}$ with input $(y,A,R)$ we have
%\begin{align}
%\inf_{\hat{T}} \sup_{\theta^*} \lv \hat{T} - \theta^* \rv_{\infty} \ge C R \lv \theta^* \rv_{2} \left(\sqrt{1 - \frac{\log(s)}{\log(r)}} \right)
%\end{align}
%where the $\inf_{\hat{T}}$ is over all estimators $\hat{T}$ and $sup_{\theta^*}$ is over $s$-sparse vectors with norm $\lv \theta^* \rv_2$.
%\end{theorem}
\begin{remark}Note that when $R  = \mathcal{O}(1/\sqrt{s})$ and $s = \mathcal{O}(\sqrt{d})$, this lower bound matches the upper bound we have for Theorem \ref{bell} (up to logarithmic factors) and hence the $\{\ell_1,\ell_2,\ell_{\infty}\}$-MU Selector is minimax optimal.\end{remark}
The proof of this theorem follows by Fano's method and is relegated to Appendix \ref{minimaxsec}.
%\subsubsection*{Previous work on Robust Sparse Regression}

\subsection{Gradient Update for the dictionary} \label{gradientsec}

We note that the update to the dictionary at each step in Algorithm \ref{Infsample} is as follows
\begin{align*}
A_{ij}^{(t)} = A_{ij}^{(t-1)} - \eta \underbrace{\left(\frac{1}{n} \sum_{k=1}^{n}\left[\sum_{p=1}^{r} \left(A_{ip}^{(t-1)} x^{k,(t)}_p x^{k,(t)}_j  - y^k_i x^{k,(t)}_j\right) \right] \right)}_{\triangleq \hat{g}_{ij}^{(t)}},
\end{align*}
for $i \in \{1,\ldots,d \}$, $j \in \{1,\ldots,r \}$ and $t \in \{1,\ldots, T \}$. If we consider the loss function at time step $t$ built using the vector samples $y^{1},\ldots, y^{n}$ and sparse estimates $x^{1,(t)},\ldots, x^{n,(t)}$,
\begin{align*}
\mathcal{L}_n(A) = \frac{1}{2n}\sum_{k=1}^n  \left \lVert y^k - A x^{k,(t)} \right  \rVert_2^2, \qquad \forall A \in \mathbb{R}^{d \times r},
\end{align*}
we can identify the update to the dictionary $\hat{g}^{(t)}$ as the gradient of this loss function evaluated at $A^{(t-1)}$,
\begin{align*}
\hat{g}^{(t)} = \frac{\partial \mathcal{L}_n(A) }{\partial A}\Big \rvert_{A^{(t-1)}}.
\end{align*}

\section{Main Results and Assumptions}
\label{mainresultssec}In this section we state our convergence result and state the assumptions under which our results are valid.
\subsection{Assumptions}
\label{assumptions}
\subsubsection*{Assumptions on $A^{*}$}
\begin{enumerate}[({A}1)]
\item
\label{assumption:incoherence}
{\bf Incoherence:} \emph{We assume the the true underlying dictionary is $\mu/\sqrt{d}$-incoherent}
\begin{align*} 
	\lvert \langle A^{*}_{i},A^{*}_{j} \rangle \rvert  & \le  \frac{\mu}{\sqrt{d}}  \quad  \forall \ i,j \in \{1,\ldots,r\}  \text{ such that, }  i\neq j.
\end{align*}
This is a standard assumption in the sparse regression literature when
support recovery is of interest. It was introduced in~\citep{fuchs,tropp} in
signal processing and independently in~\citep{zhaoyu,meins} in
statistics. We can also establish guarantees under the strictly weaker
$\ell_{\infty}$-sensitivity condition \citep[cf.][]{gautier} used
in analyzing sparse estimators under in-variable uncertainty in
\citep{belloniconic,MUimprov}. The
$\{\ell_1,\ell_2,\ell_{\infty}\}$-MU selector that we use for our
sparse recovery step also works with this more general assumption,
however for ease of exposition we assume $A^*$ to be
$\mu/\sqrt{d}$-incoherent.
\item {\bf Normalized Columns}:  \emph{We assume that all the columns of $A^{*}$ are normalized to $1$,}
\begin{align*}
\lVert A^{*}_{i} \rVert_{2} = 1 \ \ \forall \ i \in \{1,\ldots,r \}.
\end{align*}
Note that the samples $\{y^i \}_{i=1}^n$ are invariant when we scale
the columns of $A^*$ or under permutations of its columns. Thus we
restrict ourselves to dictionaries with normalized columns and label
the entire equivalence class of dictionaries with permuted columns and
varying signs as $A^*$. %We will converge linearly to the dictionary in this equivalence class that is closest to our initial estimate $A^{(0)}$.
\item {\bf Bounded max-norm:} \emph{We assume that $A^{*}$ is bounded in matrix infinity norm}
\begin{align*}
\lVert A^{*} \rVert_{\infty}  & \le \frac{C_b}{s},
 %\lVert A^{*}_i \rVert_{\infty} & > \frac{3C_b}{4s}   \ \  \forall \ i \in \{1,\ldots,r \}.
\end{align*}
\emph{where $C_b = 1/2000M^2$}. This is in contrast with previous work that imposes conditions on the operator norm of $A^*$ \citep{arora1,agarwal1,arora2}. Our assumptions help provide guarantees under alternate assumptions and it also allows the operator norm to grow with dimension, whereas earlier work requires $A^*$ to be such that $\lv A^{*} \rv_{op} \le C\left(\sqrt{r/d}\right)$. In general the infinity norm and operator norm balls are hard to compare. However, one situation where a comparison is possible is if we assume the entries of the dictionary to be drawn iid from a Gaussian distribution $\mathcal{N}(0,\sigma^2)$. Then by standard concentration theorems, for the operator norm condition to be satisfied we would need the variance ($\sigma^2$) of the distribution to scale as $\mathcal{O}(1/d)$ while, for the infinity norm condition to be satisfied we need the variance to be $\tilde{\mathcal{O}}(1/s^2)$. This means that modulo constants the variance can be much larger for the infinity norm condition to be satisfied than for the operator norm condition. %We require a lower bound on the column-wise infinity norm so that our the columns of our iterates $A^{(t)} \neq 0$,  $\forall t \in \{1,\ldots, T \}$.
\item {\bf Separation:} \emph{We assume that $\forall i \in \{1,\ldots,r\}$}
\begin{align*}
\lVert A_i^* \rVert_{\infty} > \frac{3C_b}{4s}, \text{ and, }     \min_{z\in\{-1,1\}} \lVert A^*_i - z A^*_j \rVert_{\infty}\ge \frac{3C_b}{2s} \qquad \forall \ j\neq i\in \{1,\ldots,r \}.
\end{align*}
This condition ensures that two dictionaries in the equivalence class with varying signs of columns or permutations are separated in infinity norm. The first condition ensures that for any column $A_i^*$ and $-A^*_i$ are separated  $\lVert A^*_i - (-A^*_i) \rVert_{\infty} \ge 3C_b/2s$.
\end{enumerate}
\subsubsection*{Assumption on the initial estimate and initialization}
\begin{enumerate}[({B}1)]
\item \emph{We require an initial estimate for the dictionary $A^{(0)}$ that is close in infinity norm,}
\begin{align*}
\lVert A^{(0)} - A^{*} \rVert_{\infty} \le \frac{C_{b}}{2s}.
\end{align*} 
This initialization combined with the separation condtion above ensures that the initial estimate $A^{(0)}$ is close to \emph{only one} dictionary in the equivalence class. The algorithm is going to be contractive, hence this will hold true throughout the run of the algorithm.
\end{enumerate}
%with $2C_b = C_R$; where $C_R = 1/2000M^2$.} Assuming $2C_b = C_R$ allows a fast random initialization, where we draw each entry of the initial estimate from the uniform distribution (on the interval $(-C_b/2s,C_b/2s )$). This allows us to circumvent the computationally heavy SVD/clustering step required in previous work to get an initial dictionary estimate \citep{arora2,agarwal1,arora1}. Note that we start with a random initialization and not with $A^{(0)} = 0$, as this causes our sparse estimator to fail (columns of $A$ need to be non-zero).  \textcolor{red}{need to make $C_R$ half of current value and add a separability assumption}%as this will be equally close to the entire equivalence class of dictionaries $A^*$ (with varying signs of columns and permutations). 
%A random initialization perturbs the initial solution to be closest to one of the dictionaries in the equivalence class which label as $A^*$ to which we then converge linearly.

\subsubsection*{Assumptions on $x^{*}$}
Next we assume a generative model on the $s$-sparse covariates $x^{*}$.  Here are the assumptions we make about the (unknown) distribution of  $x^*$.

\begin{enumerate}[({C}1)]
%\item {\bf Support of $x^{*}$:} We assume that subset of $s$ non-zero entries is chosen uniformly at random from sets of size $r$.
\item {\bf Conditional Independence:} \emph{We assume that distribution of non-zero entries of $x^{*}$ is conditionally independent and identically distributed.} That is, $x^{*}_{i} \ci x^{*}_{j} | x^{*}_{i}, x^{*}_{j} \neq 0$.
\item {\bf Sparsity Level:}\emph{We assume that the level of sparsity $s$ is bounded 
\begin{align*}
2 \le s \le \min(2\sqrt{d}, C_b \sqrt{d}, C\sqrt{d}/\mu),
\end{align*}
where $C$ is an appropriate global constant such that $A^*$ satisfies assumption (D3), see Remark~\ref{rem:kinfinity}.} For incoherent dictionaries, this upper bound is tight up to constant factors for sparse recovery to be feasible \citep{donoho,gribonval}.
\item {\bf Boundedness: }\emph{Conditioned on the event that $i$ is in the subset of non-zero entries, we have 
\begin{align*}
m \le  \lvert x^{*}_{i} \rvert \le M,
\end{align*}
 with $m \ge 32 R^{(0)}M(R^{(0)}(s+1) +s/\sqrt{d})$ and $M>1$}. This is needed
 for the thresholded sparse estimator to correctly predict the sign of
 the true covariate ($sgn(x) = sgn(x^*)$).  We can also relax the
 boundedness assumption: it suffices for the $x_i^*$ to have
 sub-Gaussian distributions.
\item {\bf Probability of support:} \emph{The probability of $i$ being in the support of $x^*$ is uniform over all $i \in \{1,2,\ldots,r\}$.} This translates to
\begin{align*}
\mathbb{P}r(x^{*}_{i} \neq 0) & = \frac{s}{r} \qquad \forall  \ i \in \{1,\ldots,r \}, \\
\mathbb{P}r(x^{*}_{i},x^{*}_{j} \neq 0)  & = \frac{s(s-1)}{r(r-1)} \qquad  \forall \ i\neq j\in \{1,\ldots,r \}.
\end{align*}
\item {\bf Mean and variance of variables in the support:} \emph{We assume that the non-zero random variables in the support of $x^*$ are centered and are normalized}
\begin{align*}
\mathbb{E}(x^{*}_{i}|x^{*}_{i} \neq 0) = 0, \qquad \mathbb{E}(x^{* 2}_{i} | x^{*}_{i} \neq 0) = 1.
\end{align*} 
%\item {\bf Variance of variables in the support:} \emph{We assume that the non-zero variables in the support of $x^*$ are normalized}
%\begin{align*}
%\mathbb{E}(x^{* 2}_{i} | x^{*}_{i} \neq 0) = 1.
%\end{align*}
\end{enumerate}%Finally we also require that at each round we pick the tuning coefficients $\lambda$ and $\nu$ such that condition (D5) stated in Section \ref{assbell} holds.  The reason for this condition is technical in nature and is needed to ensure that the $\{\ell_1,\ell_2,\ell_{\infty} \}$-MU Selector has an error bound with appropriate constants. 
%We note that no attempt has been made to optimize the constants in our bounds. %These constants correspond to the rate of linear convergence being $3/4$, for different convergence rates they will change appropriately. The important phenomenon presented here is the scaling with dimension of the dictionary and sparsity level. 
We note that these assumptions (A1), (A2) and (C1) - (C5) are similar
to those made in \citep{arora1, agarwal1}. \citet{agarwal1} require
the matrices to satisfy the restricted isometry property, which is
strictly weaker than $\mu/\sqrt{d}$-incoherence, however they can tolerate a
much lower level of sparsity ($d^{1/6}$). %We can also handle sub-Gaussian additive noise ($y = A^*x^* +$ noise) with its magnitude \emph{small} as compared to $A^*x^*$, however we ignore this in our analysis.

\subsection{Main Result}
\begin{theorem} \label{maintheorem}
Suppose that true dictionary $A^*$ and the distribution of the $s$-sparse samples $x^*$ satisfy the assumptions stated in Section \ref{assumptions} and we are given an estimate $A^{(0)}$ such that $\lv A^{(0)} - A^* \rv_{\infty} \le R^{(0)} \le C_b/2s$. If we are given $\{n^{(t)}\}_{t=1}^{T}$ i.i.d. samples in every iteration with $n^{(t)} = \Omega\left(\frac{r}{s (R^{(t-1)})^2}\log(dr/\delta)\right)$ then Algorithm \ref{Infsample} with parameters $(\{\tau^{(t)} \}_{t=1}^T,\{\gamma^{(t)} \}_{t=1}^T,\{\lambda^{(t)} \}_{t=1}^T,\{\nu^{(t)} \}_{t=1}^T, \eta)$ chosen as specified in Section \ref{assumptions} after $T$ iterations returns a dictionary $A^{(T)}$ such that,
\begin{align*}
\lv A^{(T)} - A^* \rv_{\infty} \le \left(\frac{7}{8}\right)^{T} R^{(0)}, \qquad \text{with probability } 1-T\delta.
\end{align*}
%where $4\eta\lvert\epsilon_n \rvert \le R^{(0)}/4$ with probability $1-\delta$.
\end{theorem}
%We note that the $\epsilon_n$ can be driven to be smaller with high probability at the cost of more samples.

\section{Proof of Convergence}
\label{convergencesec}In this section we prove the main convergence result. To prove this we analyze the gradient update to the dictionary at each step. We can decompose this gradient update (which is a random variable) into a first term which is its expected value and a second term which is its deviation from expectation. We will prove a deterministic convergence result by working with the expected value of the gradient and then appeal to standard concentration arguments to control the deviation of the gradient from its expected value with high probability. 

By Lemma \ref{lem:signbound}, Algorithm \ref{Infsample} is guaranteed to estimate the sign pattern correctly at every round of the algorithm, $sgn(x) = sgn(x^{*})$ (see proof in Appendix \ref{auxlemma}). Also note that by assumption (B1), the initial dictionary $A^{(0)}$ is close to \emph{only one} dictionary $A^*$ in the equivalence class.

To un-clutter notation let, $A^{*}_{ij} = a^{*}_{ij}$, $A^{(t)}_{ij} = a_{ij}, A^{(t+1)}_{ij} = a^{'}_{ij}$. The $k^{th}$ coordinate of the $m^{th}$ covariate is written as $x^{m*}_{k}$. Similarly let $x_k^m$ be the $k^{th}$ coordinate of the estimate of the $m^{th}$ covariate at step $t$.  Finally let $R^{(t)} = R$, $n^{(t)} = n$ and  $\hat{g}_{ij}$ be the $(i,j)^{th}$ element of the gradient with $n$ $(n^{(t)})$ samples at step $t$. Unwrapping the expression for $\hat{g}_{ij}$ we get,
\begin{align*}
\hat{g}_{ij}  &= \frac{1}{n} \sum_{m=1}^{n} \left[\sum_{k=1}^r \left(a_{ik} x_k^m x^m_j\right) - y^m_i x^m_j \right]
  = \frac{1}{n} \sum_{m=1}^{n}\left[\sum_{k=1}^r \Big( a_{ik} x_k^{m} - a_{ik}^{*} x_k^{m*}\Big)x_j^{m}\right]\\
& = \mathbb{E} \left[ \sum_{k=1}^r \Big( a_{ik} x_k - a_{ik}^{*} x_k^{*}\Big)x_j\right] \\ & \quad + \left[ \frac{1}{n} \sum_{m=1}^{n}\left[ \sum_{k=1}^r \Big( a_{ik} x_k^{m} - a_{ik}^{*} x_k^{m*}\Big)x_j^{m} \right] - \mathbb{E} \left[ \sum_{k=1}^r \Big( a_{ik} x_k - a_{ik}^{*} x_k^{*}\Big)x_j \right] \right] \\
& = g_{ij} + \underbrace{\hat{g}_{ij} - g_{ij}}_{\triangleq \epsilon_n},
\end{align*}
where $g_{ij}$ denotes $(i,j)^{th}$ element of the expected value (infinite samples) of the gradient. The second term $\epsilon_n$ is the deviation of the gradient from its expected value. By Theorem \ref{finite} we can control the deviation of the sample gradient from its mean via an application of McDiarmid's inequality. With this notation in place we are now ready to prove Theorem \ref{maintheorem}.

\begin{Proof}[Proof of Theorem \ref{maintheorem}] First we analyze the structure of the expected value of the gradient.

\emph{Step 1}: Unwrapping the expected value of the gradient we find it decomposes into three terms 
\begin{multline*} 
g_{ij}  = \mathbb{E} \left(a_{ij}x^{2}_j - a_{ij}^{*} x^{*}_j x_j \right) + \mathbb{E} \left[ \sum_{k \neq j} a_{ik} x_k x_j - a^{*}_{ik} x_{k}^{*} x_j \right]\\
 = \underbrace{(a_{ij} - a_{ij}^{*}) \frac{s}{r} \mathbb{E} \left[ x_j^{2} | x^*_j \neq 0 \right]}_{\triangleq g^c_{ij}} + \underbrace{a_{ij}^{*} \frac{s}{r} \mathbb{E} \left[ (x_j - x_j^*)x_j | x^*_j \neq 0 \right]}_{\triangleq\Xi_1} +  \underbrace{\mathbb{E} \left[ \sum_{k \neq j} a_{ik} x_k x_j - a^{*}_{ik} x_{k}^{*} x_j \right]}_{\triangleq\Xi_2}.
\end{multline*}
The first term $g^{c}_{ij}$ points in the \emph{correct direction} and
will be useful in converging to the right answer. The other terms
could be in a bad direction and we will control their magnitude with
Lemma~\ref{Xicont} such that $\lvert \Xi_1 \rvert + \lvert \Xi_2
\rvert \le \frac{s}{3r} R$. The proof of Lemma \ref{Xicont} is the
main technical challenge in the convergence analysis to control the
error in the gradient. Its proof is deferred to the appendix.

\emph{Step 2}: Given this bound, we analyze the gradient update,
\begin{align*}
a_{ij}^{'} & = a_{ij} - \eta \hat{g}_{ij}  = a_{ij} - \eta (g_{ij} + \epsilon_n) 
 = a_{ij} - \eta \left[g^{c}_{ij} + (\Xi_1 + \Xi_2) + \epsilon_n \right].
\end{align*}
So if we look at the distance to the optimum $a^*_{ij}$ we have the relation,
\begin{align*}
a^{'}_{ij} - a^{*}_{ij} & = a_{ij} - a^{*}_{ij} - \eta (a_{ij} - a_{ij}^{*}) \frac{s}{r} \mathbb{E} \left[ x_j^{2} | x^*_j \neq 0 \right] - \eta \left\{ (\Xi_1 + \Xi_2) + \epsilon_n \right\}. 
\end{align*}
Taking absolute values, we get
\begin{align*}
\lvert a^{'}_{ij} - a^{*}_{ij} \rvert & \overset{(i)}{\le} \left(1 -
\eta \frac{s}{r}\mathbb{E}\left[x_{j}^{2} | x^*_{j} \neq 0 \right] \right)\lvert a_{ij} - a^{*}_{ij} \rvert + \eta \left\{\lvert \Xi_1\rvert + \lvert \Xi_2 \rvert + \lvert \epsilon_n \rvert \right\}\\
  & \overset{(ii)}{\le} \left(1 - \eta
  \frac{s}{r}\mathbb{E}\left[x_{j}^{2} | x^*_{j} \neq 0 \right] \right)\lvert a_{ij} - a^{*}_{ij} \rvert + \eta \left(\frac{s}{3r} R \right) + \eta\lvert \epsilon_{n}\rvert\\
 & \le \left(1 - \eta \frac{s}{r}\left\{\mathbb{E}\left[x_{j}^{2} |
 x^*_{j} \neq 0 \right] -\frac{1}{3} \right\}\right)R + \eta \lvert \epsilon_n \rvert,
\end{align*}
provided the first term is at non-negative.
Here, $(i)$ follows by triangle inequality and $(ii)$ is by Lemma
\ref{Xicont}. Next we give an upper and lower bound on
$\mathbb{E}\left[x_{j}^{2} | x^*_{j} \neq 0 \right]$. We would expect
that as $R$ gets smaller this variance term approaches
$\mathbb{E}\left[x_{j}^{*2} | x^*_{j} \neq 0 \right] = 1$. By invoking
Lemma \ref{varcont} we can bound this term to be $\frac{2}{3} \le
\mathbb{E}\left[x_{j}^{2} | x^*_{j} \neq 0 \right] \le \frac{4}{3}$.
We want the first term to contract at a rate $3/4$; it suffices to have
\begin{align*}
0 \overset{(i)}{\le}  \left(1 - \eta
\frac{s}{r}\left\{\mathbb{E}\left[x_{j}^{2} | x^*_{j} \neq 0 \right] -\frac{1}{3} \right\}\right) \overset{(ii)}{\le} \frac{3}{4}.
\end{align*}
Coupled with Lemma \ref{varcont}, Inequality~$(i)$ follows from
$\eta \le \frac{r}{s}$ while $(ii)$ follows from $\eta \ge \frac{3r}{4s}$. We also have by Theorem \ref{finite} that $\eta \lvert \epsilon_n \rvert \le R/8$ with probability $1-\delta$. So if we unroll the bound for
$t$ steps we have,
\begin{align*}
\lvert a^{(t)}_{ij} - a^*_{ij} \rvert & \le \frac{3}{4} R^{(t-1)}  + \eta \lvert \epsilon_n \rvert \le \frac{3}{4}R^{(t-1)} + \frac{1}{8}R^{(t-1)} = \frac{7}{8}R^{(t-1)} \le \left(\frac{7}{8}\right)^{t} R^{(0)}.
%& \le  \frac{3}{4}\left(\frac{3}{4} R^{(t-2)}   + \eta \lvert \epsilon_n \rvert \right) + \eta \lvert \epsilon_n \rvert\\\
%& \le \left( \frac{3}{4} \right)^{t}  R^{(0)}  + \left\{ \sum_{q=0}^{t-1} \left(\frac{3}{4}\right) ^{q}\right\} \left(\eta \lvert\epsilon_n \rvert\right)\\
%& \le \left( \frac{3}{4} \right)^{t}  R^{(0)}   + 4 \eta \lvert \epsilon_n \rvert & \text{as} \left( \sum_{q=0}^{\infty} (3/4) ^{q} = 4 \right).
\end{align*}
We also have $\eta \rvert \epsilon_n\rvert \le R/8 \le R^{(0)}/8$ with probability at least $1-\delta$ in each iteration, for all $t \in \{1,\ldots, T\}$; thus by taking a union bound over the iterations we are guaranteed to remain in our initial ball of radius $R^{(0)}$ with high probability, completing the proof.
\end{Proof}

\section{Conclusion}
%In this paper we present an alternating minimization algorithm for dictionary learning and identify conditions under which random initialization works. We can also handle sparsity scaling as $\sqrt{d}$ which is information theoretically optimal for incoherent dictionaries while allowing for an arbitrary level of overcompleteness. We  analyze the error in the recovering the dictionaries in infinity norm that allows for sharper control than previous work that analyze error the maximum column-wise error.

An interesting question would be to further explore and analyze the
range of algorithms for which alternating minimization works and
identifying the conditions under which they provably converge. Going beyond sparsity
$\sqrt{d}$ still remains challenging, and as noted in previous work
alternating minimization also appears to break down experimentally and
new algorithms are required in this regime. Also all theoretical work
on analyzing alternating minimization for dictionary learning seems to
rely on identifying the signs of the samples ($x^*$) correctly at
every step. It would be an interesting theoretical question to analyze
if this is a necessary condition or if an alternate proof strategy and
consequently a bigger radius of convergence are possible. Lastly, it
is not known what the optimal sample complexity for this problem is
and lower bounds there could be useful in designing more sample
efficient algorithms.
\subsubsection*{Acknowledgments}
We gratefully acknowledge the support of the NSF through grant
IIS-1619362, and of the Australian
Research Council through an Australian Laureate Fellowship
(FL110100281) and through
the ARC Centre of Excellence for Mathematical and Statistical
Frontiers. Thanks also to the Simons
Institute for the Theory of Computing Spring 2017 Program on
Foundations of Machine Learning. 

The authors
would like to thank Sahand Negahban for pointing out an error in the $\mu$-incoherence assumption in an
earlier version. The authors would like to thank Shivam Garg for pointing us to an error in the claim about random initialization in a previous version of this paper.

\nocite{*}
\bibliography{ref} 
\appendix
\newpage
\section{Additional details  for the proof of convergence}
For Appendix~\ref{auxlemma} and~\ref{finiteapp}, we borrow the
notation from Section~\ref{convergencesec}. In Appendix~\ref{auxlemma}
we prove Lemma~\ref{maxlemma} that controls an error term which will
be useful in establishing Lemma~\ref{Xicont} that bounds the error
terms in the gradient, $\Xi_1$ and $\Xi_2$. Corollary~\ref{usefulbell}
establishes the error bound for the sparse estimate while Lemma~\ref{lem:signbound} establishes that the sparse estimate after the
thresholding step has the correct sign. In Appendix~\ref{finiteapp}, we
establish finite sample guarantees.
\subsection{Proof of Auxillary Lemmas}\label{auxlemma}
Before we prove Lemma~\ref{Xicont}, which controls the terms in the
gradient, we prove Lemma~\ref{maxlemma}, which
will be vital in controlling the cross-term in the gradient.
\begin{lemma}\label{maxlemma} Let the assumptions stated in Section \ref{assumptions} hold. Then at each iteration step we have the guarantee that
\begin{align*}
 \left \lvert \max_{k: k\neq j}\left\{\mathbb{E}\left[a_{ik}x_k x_j - a^*_{ik}x^*_k x_j \lvert x^*_k \neq 0, x^*_j \neq 0 \right]\right\}\right \rvert \le \frac{R}{6(s-1)}.
\end{align*}
\end{lemma}
\begin{Proof}
Let us define
\begin{align*}
\Gamma \triangleq  \max_{k: k\neq j}\left\{\mathbb{E}\left[a_{ik}x_k x_j - a^*_{ik}x^*_k x_j \lvert x^*_k \neq 0, x^*_j \neq 0 \right]\right\},
\end{align*}
and let us define the event $\mathcal{E}_{jk} \triangleq \{x^*_j \neq 0, x^*_k \neq 0 \}$. Expanding $\Gamma$,
\begin{align*}
\Gamma & = \max_{k: k\neq j} \Big\{ \mathbb{E}\left[a_{ik}(x_k - x^*_k + x^*_k)(x_j - x^*_j + x^*_j) -a^*_{ik}x^*_k(x_j - x^*_j + x^*_j) \lvert \mathcal{E}_{jk} \right] \Big\}\\
& = \max_{k: k\neq j} \Big\{  \underbrace{(a_{ik}-a_{ik}^*) \mathbb{E} \left[x_k^*(x_j-x_j^*) \lvert \mathcal{E}_{jk} \right]}_{\triangleq n_1} + \underbrace{a_{ik} \mathbb{E} \left[(x_k - x_k^*)x_j^* \lvert \mathcal{E}_{jk} \right]}_{\triangleq n_2}\\ & \qquad \qquad \qquad \qquad \qquad  + \underbrace{a_{ik} \mathbb{E} \left[(x_k - x_k^*)(x_j-x_j^*) \lvert \mathcal{E}_{jk} \right]}_{\triangleq n_3}+ \underbrace{(a_{ik}-a^*_{ik})\mathbb{E} \left[ x_k^*x_j^* \lvert \mathcal{E}_{jk} \right]}_{\triangleq n_4}\Big\}.
\end{align*}
Given that the non-zero entries of $x^*$ are independent and mean zero we have $n_4 = 0$. Next we see $n_1,n_2$ and $n_3$ are bounded above as
\begin{align*}
n_1 & \le \lvert a_{ik}- a^*_{ik}\rvert M  \lv  x- x^* \rv_{\infty} \le R M  \lv  x- x^* \rv_{\infty}\\
n_2 & \le  \lvert a_{ik}\rvert M  \lv  x- x^* \rv_{\infty} \le (\lvert a_{ik}^*\rvert + R) M  \lv  x- x^* \rv_{\infty} \\
n_3 & \le \lvert a_{ik}\rvert\lv x- x^* \rv_{\infty}^2 \le (R + \lvert a^*_{ik} \rvert)\lv x- x^* \rv_{\infty}^2,
\end{align*}
these follow as $\lvert x^*_j \rvert \le M$, $\lvert x_j - x^*_j\rvert \le \lv x - x^* \rv_{\infty}$ and $\lvert a_{ik} - a^*_{ik} \rvert \le R$. The goal now is to show that $n_1 \le R/30(s-1)$, $n_2 \le R/15(s-1)$ and $n_3 \le R/15(s-1)$. Let us unwrap the first term of $n_1$
\begin{align}
\nonumber n_1 & \le R M \lv x -x^* \rv_{\infty} \overset{(i)}{\le} \frac{R}{30(s-1)}\left[30(s-1)M\cdot 16 R M \left(R(s+1) + \frac{s}{\sqrt{d}} \right) \right]\\
\nonumber & \overset{(ii)}{\le} \frac{R}{30(s-1)}\left[ 30 (s-1)M \cdot \frac{8 C_b M}{s} \left( \frac{C_b(s+1)}{2s} +2 \right)\right]\\
\nonumber & =\frac{R}{30(s-1)}\left[ 240 M^2 \underbrace{\left(\frac{s-1}{s}\right)}_{\le 1} C_b \left(C_b \underbrace{\left(\frac{(s+1)}{2s}\right)}_{\le 3/4} + 2 \right) \right]\\
& \le \frac{R}{30(s-1)} \underbrace{\left[ 240 M^2  C_b \left(\frac{3}{4}C_b + 2\right)\right]}_{\triangleq \xi_1}, \label{1cond}
\end{align} 
where $(i)$ follows by invoking Corollary \ref{usefulbell} and $(ii)$
follows as $s\le 2 \sqrt{d}$ and $R \le C_b/2s$. Our choice $C_b = 1/2000M^2$ ensures
that $\xi_1 \le 1$. The second term in the upper bound on $n_2$ can be bounded by the same technique as we
used to bound $n_1$, giving $RM\lv x-x^* \rv_{\infty} \le R/30(s-1)$.
For the first term in $n_2$, we have
\begin{align}
\nonumber \lvert a^*_{ik} \rvert M \lv x-x^* \rv_{\infty} & \le \frac{R}{30(s-1)}\left[ 480 \frac{(s-1)}{s} M^2 C_b \left(R(s+1) + \frac{s}{\sqrt{d}} \right)\right]\\
& \le \frac{R}{30(s-1)} \underbrace{\left[ 480 M^2 C_b \left( C_b \frac{(s+1)}{2s} +2 \right)\right]}_{\triangleq \xi_2}, \label{2cond}
\end{align}
where these inequalities follow by invoking Corollary~\ref{usefulbell} and by the upper bounds on $\lvert a^*_{ik}\rvert$ and $R$. Again our choice $C_b = 1/2000M^2$ ensures that $\xi_2 \le 1$ which leaves us with the upper bound on $n_2 \le \frac{R}{15(s-1)}$. Finally to bound $n_3$ we observe that the first term is bounded as follows,
\begin{align*}
R \lv x -x^* \rv_{\infty}^2 & \le \frac{R}{30(s-1)} \left[
30(s-1)\cdot 16^2 R^2 M^2 \left(R(s+1) + \frac{s}{\sqrt{d}} \right)^2\right]\\
& \le \frac{R}{30(s-1)} \left[\sqrt{30(s-1)}\cdot 16 \frac{C_b}{2s} M \left(C_b\frac{(s+1)}{2s} + 2 \right) \right]^2\\
& \le \frac{R}{30(s-1)},
\end{align*}
where the last inequality is due to the fact that $\xi_1 \le 1$.  We have $\lvert a_{ik}^*\rvert \le C_b/s$ and similar arguments as above can be used to show that the second term in $n_3$ is also bounded above by $\frac{R}{30(s-1)}$. Having controlled $n_1,n_2$ and $n_3$ at the appropriate levels completes the proof and yields the desired bound on $\Gamma$.
\end{Proof}
\begin{lemma} \label{Xicont}Let the assumptions stated in Section \ref{assumptions} hold. Then at each iteration step we can bound the error terms in the gradient as
\begin{align*}
\lvert \Xi_1 \rvert & = \left\lvert a_{ij}^{*} \frac{s}{r} \mathbb{E} \left[ (x_j - x_j^*)x_j | x^*_j \neq 0 \right] \right\rvert \le \frac{s}{6r} R\\
\lvert \Xi_2 \rvert & = \left\lvert \mathbb{E} \left[ \sum_{k \neq j} a_{ik} x_k x_j - a^{*}_{ik} x_{k}^{*} x_j \right] \right\rvert  \le  \frac{s}{6r} R.
\end{align*}
\end{lemma}
\begin{Proof}
\emph{Part 1}-We first prove the bound on $\Xi_1$. We start by unpacking $\Xi_1$
\begin{align}
\lvert \Xi_1 \rvert & = \left\lvert \frac{s}{r} a_{ij}^{*}  \mathbb{E} \left[ (x_j - x_j^*)x_j | x^*_j \neq 0 \right] \right\rvert \nonumber\\
& \le \frac{s}{r} \lvert a^*_{ij} \rvert \cdot \left\lvert \mathbb{E} \left[ (x_j - x_j^*)(x^*_j + x_j-x_j^* ) | x^*_j \neq 0 \right]\right \rvert \nonumber\\
& \overset{(i)}{\le} \frac{s}{r}  \lvert a^*_{ij} \rvert M \cdot
\mathbb{E}\left[ \lv x- x^* \rv_{\infty}| x^*_j \neq 0 \right] +
\frac{s}{r} \lvert a^*_{ij} \rvert \mathbb{E}\left[ \lv x-x^* \rv_{\infty}^2| x^*_j
\neq 0\right] \nonumber\\
& \overset{(ii)}{\le} \frac{s}{r} \lvert a^*_{ij} \rvert M\cdot \left(16 R M \left( (s+1)R + \frac{s}{\sqrt{d}}\right)\right)+ \frac{s}{r}\lvert a^*_{ij} \rvert  \left(16 R M \left( (s+1)R + \frac{s}{\sqrt{d}}\right)\right)^2 \nonumber\\
& = \frac{s}{6r} R \left\{ 96 \lvert a^*_{ij} \rvert M^2  \left(R(s+1) + \frac{s}{\sqrt{d}}\right)+ 6 \lvert a^*_{ij}\rvert R \left(16 M\left( R(s+1)+\frac{s}{\sqrt{d}}\right) \right)^2\right\} \label{curlyequation} \\
& \le \frac{s}{6r}R, \nonumber
\end{align}
where $(i)$ follows by triangle inequality and $\lvert x^*_{j} \rvert \le M$ and, $(ii)$ follows by Corollary \ref{usefulbell}. It can be shown that in \eqref{curlyequation} the term in the curly braces is $\le 1$ by arguments similar to those used in Lemma \ref{maxlemma}  (because $R \le 1/4000M^2 s$, $s \le 2\sqrt{d}$ and $\lvert a^*_{ij} \rvert \le 1/2000M^2 s$), thus establishing the desired bound on $\lvert \Xi_1 \rvert$ .

\emph{Part 2}- Expanding $\Xi_2$ we find
\begin{align*}
\lvert \Xi_2 \rvert & = \left\lvert \mathbb{E} \left[ \sum_{k \neq j} a_{ik} x_k x_j - a^{*}_{ik} x_{k}^{*} x_j \right] \right\rvert \\
& \overset{(i)}{=} \frac{s(s-1)}{r(r-1)} \left\lvert \mathbb{E} \left[\sum_{k\neq j}  a_{ik}x_k x_j  - a^*_{ik}x^*_k x_j \lvert x^*_k \neq 0, x^*_j \neq 0\right]\right\rvert\\
& \le \frac{s(s-1)}{r(r-1)}\cdot (r-1) \left \lvert \max_{k\neq j}\left\{\mathbb{E}\left[a_{ik}x_k x_j - a^*_{ik}x^*_k x_j \lvert x^*_k \neq 0, x^*_j \neq 0 \right]\right\}\right \rvert\\
%&\le \frac{s(s-1)}{r} \left \lvert \max_{k: k\neq j}\left\{\mathbb{E}\left[a_{ik}x_k x_j - a^*_{ik}x^*_k x_j \lvert x^*_k \neq 0, x^*_j \neq 0 \right]\right\}\right \rvert\\
& = \frac{s}{6r} R \left(\frac{6(s-1)}{R} \left \lvert \max_{k\neq j}\left\{\mathbb{E}\left[a_{ik}x_k x_j - a^*_{ik}x^*_k x_j \lvert x^*_k \neq 0, x^*_j \neq 0 \right]\right\}\right \rvert \right)\\
& \overset{(ii)}{\le} \frac{s}{6r} R,
\end{align*}
where $(i)$ follows from assumption (C4) and $(ii)$ follows by invoking Lemma \ref{maxlemma}.
\end{Proof}

\begin{lemma}\label{varcont} Let the assumptions stated in Section \ref{assumptions} hold. Then at each iteration step we can bound the variance of the estimate,
\begin{align*}
\frac{2}{3} \le \mathbb{E}\left[x_{j}^{2} | x^*_{j} \neq 0 \right] \le \frac{4}{3}.
\end{align*}
\end{lemma}
\begin{Proof} Consider the expectation of the random variable $x_j^2 -
x_j^{*2}| x^*_{j} \neq 0$. We have
\begin{align*}
x_{j}^{2} - x^{*2}_{j} & \le \lvert x_j + x^*_j \rvert \lv x - x^* \rv_{\infty}\\
& = \lvert 2x^*_j + x_j - x^*_j \rvert \lv x-x^* \rv_{\infty} \le 2 \lvert x^*_j \rvert \lv x - x^* \rv_{\infty} + \lv x-x^* \rv_{\infty}^2\\
& \le \underbrace{2 M \lv x- x^* \rv_{\infty} + \lv x - x^* \rv_{\infty}^2}_{\triangleq \xi_3}.
\end{align*}
Note that $\xi_3 \le \frac{1}{3}$, if $\lv x - x^* \rv_{\infty} \le \frac{1}{3}\left( \sqrt{3} \sqrt{3M^2 + 1} - 3M\right)$. We also have an upper bound on $\lv x- x^* \rv_{\infty}$ by Corollary \ref{usefulbell}
\begin{align*}
\lv x - x^* \rv_{\infty} & \le 16 R M \left( R(s+1) + \frac{s}{\sqrt{d}}\right) \le \underbrace{\frac{8}{s}}_{\le 4} C_b M \left( C_b \underbrace{\frac{s+1}{2s}}_{\le 3/4} + \underbrace{\frac{s}{\sqrt{d}}}_{\le 2} \right)\\
& \le 4 C_b M\left( \frac{3}{4} C_b + 2 \right).
\end{align*}
 Our choice $C_b = 1/2000M^2$ with $M>1$ guarantees that
\begin{align}
4 C_b M\left( \frac{3}{4} C_b + 2 \right) \le \frac{1}{3}\left( \sqrt{3} \sqrt{3M^2 + 1} - 3M\right), \label{3cond}
\end{align}
this yields the claimed bound.
\end{Proof}

The next corollary establishes an infinity norm bound on the error in the sparse estimate under the assumptions made in Section \ref{assumptions} and choice of parameters in Section \ref{algosec}.
\begin{corollary}\label{usefulbell1}
Under the assumptions specified in Section \ref{assumptions} and choice of parameters for Algorithm \ref{Infsample} in Section \ref{algosec} we have the bound for all $t \in \{1,\ldots,T\}$ and $k \in \{1,\ldots,n\}$,
\begin{align*}
\lv w^{k,(t)} - x^{k*} \rv_{\infty} & \le 16 R^{(t-1)} M \left( R^{(t-1)} (s+1) + \frac{s}{\sqrt{d}} \right),
\end{align*}
where $w^{k,(t)}$ is as defined in Algorithm \ref{Infsample}.
\end{corollary}
\begin{Proof} We have $\lv x^{k*} \rv_2 \le \sqrt{s}M$, $\lv x^{k*} \rv_{\infty} \le M$ thus plugging this into Theorem \ref{bell} gives us the desired result.
\end{Proof}

The next theorem guarantees that at each round of the algorithm, under the assumptions stated in Section \ref{assumptions}, we correctly predict the sign pattern. 

\begin{lemma}\label{lem:signbound} Under the assumptions (A1)-(A6),(B1) and (C1)-(C5) stated in Section \ref{assumptions} with
\begin{align} \label{as:mbound}
32 R^{(0)}M\left(R^{(0)}(s+1)+\frac{s}{\sqrt{d}}\right) < m,  
\end{align}
and under the choice of the parameters $\eta,R^{(t)},\tau^{(t)},\gamma^{(t)},\lambda^{(t)}$ and $\nu^{(t)}$ specified in Section \ref{algosec} for all $t \in \{1,2,\ldots, T\}$ we have the guarantee that Algorithm \ref{Infsample} returns a sparse estimate $\{x^{k,(t)}\}_{k=1}^n$ such that,
\begin{align*}
sgn(x^{k,(t)}) & = sgn(x^{k*}), && \forall k \in \{ 1,2,\ldots,n \}.
\end{align*}
\end{lemma}
\begin{Proof} Under the assumptions stated we can invoke Corollary \ref{usefulbell1} to get,
\begin{align}\label{eq:infbound}
\lv	w^{k,(t)} - x^{k*} \rv_{\infty} & \le 16 R^{(t-1)}M \left( R^{(t-1)}(s+1) + \frac{s}{\sqrt{d}} \right) && \forall k \in \{1,\ldots, n \},t\in \{1,\ldots,T\},
\end{align}
where $w^{k,(t)}$ is defined as in Algorithm \ref{Infsample}. Note that the thresholds are defined by the schedule,
\begin{align*}
\tau^{(t)} = 16 R^{(t-1)}M \left(R^{(t-1)}(s+1)+\frac{s}{\sqrt{d}} \right).
\end{align*}
By definition $x^{k,(t)}$ is the coordinate-wise thresholded estimate,
\begin{align*}
x^{k,(t)}_{l} & = w^{k,(t)}_l \mathbb{I}\left(\lvert w_l^{k,(t)} \rvert > \tau^{(t)} \right) && \forall l \in \{1,2,\ldots,r \}.
\end{align*}
We know that for all $t>1$ we have $R^{(t)} < R^{(0)}$. So by the
infinity norm bound in the above display \eqref{eq:infbound} and, by
the assumptions on the distribution of $x^*$, we have that
\begin{align*}
sgn\left(x^{k,(t)}\right) & = sgn\left(x^{k*}\right) && \forall k \in \{ 1,2,\ldots,n \}.
\end{align*}
This follows as the thresholding step only zeros out the non-zero elements in $x^{k,(t)}$ that are not in $supp(x^{k*})$.
\end{Proof}

\begin{corollary}\label{usefulbell}
Under the assumptions specified in Section \ref{assumptions} and choice of parameters for Algorithm \ref{Infsample} in Section \ref{algosec} we have the bound for all $t \in \{1,\ldots,T\}$ and $k \in \{1,\ldots,n\}$,
\begin{align*}
\lv x^{k,(t)} - x^{k*} \rv_{\infty} & \le 16 R^{(t-1)} M \left( R^{(t-1)} (s+1) + \frac{s}{\sqrt{d}} \right),
\end{align*}
where $x^{k,(t)}$ is as defined in Algorithm \ref{Infsample}.
\end{corollary}
\begin{Proof} Note that by Lemma \ref{lem:signbound} we have that $sgn(x^{k,(t)})=sgn(x^{k*})$.  Thus for any $l \in \{1,\ldots,r\}$ if $l \notin supp(x^{k*})$ then the choice of threshold of $\tau^{(t)} = 16 R^{(t-1)}M \left(R^{(t-1)}(s+1)+\frac{s}{\sqrt{d}} \right)$ implies that,
\begin{align*}
\lvert x^{k,(t)}_{l} - x^{k*}_l \rvert  &=\lvert x^{k,(t)}_{l}\rvert  \le 16 R^{(t-1)} M \left( R^{(t-1)} (s+1) + \frac{s}{\sqrt{d}} \right).
\end{align*}
While for $l \in supp(x^{k^*})$ Corollary \ref{usefulbell1} implies
\begin{align*}
\lvert x^{k,(t)}_{l} - x^{k*}_l \rvert & \le 16 R^{(t-1)} M \left( R^{(t-1)} (s+1) + \frac{s}{\sqrt{d}} \right).
\end{align*}
This completes the proof.
\end{Proof}
\subsection{Finite Sample Guarantees}\label{finiteapp}
 
In this section, we establish finite sample guarantees and state convergence results used in the proof of convergence of our algorithm.
\begin{theorem}\label{finite} Let $\epsilon_{n} \le \frac{R}{8\eta}$, where $\frac{3r}{4s}\le\eta\le \frac{r}{s}$ is the step-size used at each gradient step. If we are given $n$ i.i.d. samples at each round where $n =
\Omega(\frac{r}{s R^2}\log(dr/\delta))$, then we have the guarantee that
\begin{align*}
\max_{i \in \{1,\ldots,r\},j \in \{1,\ldots,d \}} \{\lvert \hat{g}_{ij} - g_{ij} \rvert \} \le \epsilon_n,
\end{align*}
with probability $1-\delta$.
\end{theorem}
\begin{Proof} We define the set $W = \{m : j \in supp(x^{m*})  \}$ and then we have that
\begin{align*}
\hat{g}_{ij} = \frac{\lvert W \rvert}{n} \cdot \underbrace{\frac{1}{\lvert W \rvert} \sum_{m \in W} \left(\sum_{k}a_{ik}x_k^m-a^*_{ik}x^{m*}_{k} \right)x^m_j}_{\triangleq\hat{g}^{W}_{ij}}.
\end{align*}
Let $x^{l*}$ be a sample such that $l \in W$. We will bound the term $\Lambda = \lvert\left(\sum_{k} a_{ik} x_{k}^{l} - a^*_{ik}x^{l*}_k\right) x^l_j\rvert$ and later invoke McDiarmid's inequality. To ease notation we drop the superscript $l$. Expanding  $\Lambda$ we get
\begin{multline*}
\Lambda  = \Bigg\lvert \sum_{k=1}^{r} (a_{ik} - a_{ik}^*)(x_k - x_k^*)(x_j-x_j^*) + (a_{ik} - a_{ik}^*)(x_k-x_k^* )x_j^* \\ + (a_{ik}-a_{ik}^*)x_k^*(x_j-x_j^*) + (a_{ik} - a_{ik}^*)x_k^* x_j^* - a_{ik}^* x_k^* (x_j - x_j^*) - a_{ik}^* x_k^* x_j^*\Bigg\rvert
\end{multline*}
Recall that by Lemma  \ref{lem:signbound} we have that $sgn(x^l) = sgn(x^{l*}) $, and $x^{l*}$ is $s$-sparse thus only $s$ terms in the above sum are non-zero. We repeatedly use the bounds,
\begin{enumerate}
\item $\lvert a_{ik}^* \rvert \le \frac{C_b}{s}$.
\item $\lvert a_{ik} - a_{ik}^* \rvert \le R \le R^{(0)} \le \frac{C_b}{2s}$.
\item $\lVert x - x^* \rVert_{\infty} \le 16 R M \left(R(s+1) + \frac{s}{\sqrt{d}} \right)$.
\item $2\le s \le 2\sqrt{d}$.
\end{enumerate}
Using these we can upper bound $\Lambda$ by 
\begin{multline*}
\Lambda  \le \frac{3C_b  M^2}{4} + \frac{10 C_b^2 M^2}{s} \left(\frac{C_b(s+1)}{2s} + 2 \right) + \frac{C_b}{2} \left(\frac{8 C_b M}{s}\left(\frac{C_b(s+1)}{2s} + 2 \right) \right)^2.
\end{multline*}
 By our choice of $C_b = 1/2000M^2$, where $M>1$ we have that 
\begin{align*}
\Lambda \le B,
\end{align*}
for an appropriate global constant $B$ (independent of $s$ and $M$). 

By simple concentration arguments we can get that $\lvert W \rvert/n$ is close to $s/r$. Conditioned on a value of $\lvert W \rvert $ by invoking McDiarmid's inequality
(Theorem~\ref{mcdiar}), we have that $\lvert \hat{g}^{W}_{ij} - \mathbb{E}\left[\hat{g}_{ij}\lvert j \in supp(x^*) \right] \rvert\le \epsilon_{W,n}$ with probability $1-2e^{-2\lvert W \rvert \epsilon_{W,n}^2/B^2}$. We demand
\begin{align}\label{finitecrux}
\epsilon_{W,n} = \frac{C \cdot r\cdot  R}{8 s \eta}, %\le \mathcal{O}\left(\frac{1}{s}\right),
\end{align}
with probability $1-c\delta/dr$ for every $(i,j)$, where $c$ and $C$
are appropriate constants such that $\lvert \hat{g}_{ij} - g_{ij} \rvert \le R/8\eta $ with probability at least $1-\delta/dr$. Thus we need $\lvert W \rvert = \Omega((\frac{s\eta}{r R})^2\log(dr/\delta))$. As $\eta$ is proportional to
$r/s$, this implies that for
\eqref{finitecrux} to hold,  we need that $\lvert W
\rvert=\Omega(1/R^2 \log(dr/\delta))$.

As stated above we have that $\lvert W \rvert/n$ is close to $s/r$ so if $\lvert W \rvert = \Omega(1/R^2 \log(dr/\delta))$ it  suffices to have $n =
\Omega(\frac{r}{s R^2}\log(dr/\delta))$. We finish the proof by a union bound over all entries of the matrix.
%over all entries of the matrix $(i,j)$. The columns of $A^*$ are normalized, so for any column, $A_i$ and $-A_i$ we have,
%\begin{align*}
%\lVert A^*_i - (-A^*_i) \rVert_{\infty} \ge \frac{1}{\sqrt{d}} \lVert A^*_i + A^*_i \rVert_2 = \frac{2}{\sqrt{d}}.
%\end{align*}
%Also far any $i\neq j$, we have
%\begin{align*}
%\lVert A^*_i - A^*_j \rVert_{\infty} \ge \frac{1}{\sqrt{d}} \lVert A^*_i - A^*_j \rVert_2 \ge \frac{1}{\sqrt{d}} \sqrt{2\left(1-\frac{\mu}{\sqrt{d}}\right)}
%\end{align*}
%where the last inequality follows as $A^*$ is $\mu$-incoherent. Thus is $\epsilon_f \le C/\sqrt{d}$ where $C$ is an appropriate constant then the stochastic noise in the gradient does not perturb the solution to another element of the equivalence class of $A^*$ (dictionary is unidentifiable up to permutations and sign flips of columns.)
\end{Proof}
\subsection{Concentration Theorems}
We recall McDiarmid's inequality \citep{mcd}.
\begin{theorem}\label{mcdiar}
Let $X_1,\ldots,X_m$ be independent random variables all taking values in the set $\mathcal{X}$. Further, let $f: \mathcal{X}^m \mapsto \mathbb{R}$ be a function of $X_1, X_2,\ldots,X_m$ that satisfies $\forall i$, $\forall x_1,\ldots,x_m,x'_i \in \mathcal{X}$,
\begin{align*}
\lvert f(x_1,\ldots,x_i,\ldots,x_m) - f(x_1,\ldots,x'_i,\ldots,x_m)\rvert \le c_i.
\end{align*}
Then for all $\epsilon >0$,
\begin{align*}
\mathbb{P}(f - \mathbb{E}\left[f\right] \ge \epsilon) \le \exp\left( \frac{-2\epsilon^2}{\sum_{i=1}^{m}c_i^2}\right).
\end{align*}
\end{theorem}
Next we present a concentration theorem for a sum of the squares of $d$ independent Gaussian random variables each with variance $\sigma^2$ ($\chi^2$-concentration theorem).
\begin{theorem}[Gaussian concentration inequality, see Theorem 5.6 in
\citep{bouch}] \label{t:lipsch} Let $X = (X_1,\ldots,X_n)$ be a vector
of $n$ independent standard normal random variables. Let $f:
\mathbb{R}^n \mapsto \mathbb{R}$ denote an $L$-Lipschitz function with
respect to Euclidean distance. Then, for all $t>0$,
\begin{align*}
\mathbb{P}(f(X) - \mathbb{E}(f(X))\ge t) & \le e^{-t^2/(2L^2)}.
\end{align*}
\end{theorem}
\begin{lemma} \label{chicon} If $\{Z_k\}_{k=1}^d \sim \mathcal{N}(0,1)$ are i.i.d. standard normal variables, then $Y \triangleq  \sigma^2 \sum_{k=1}^d Z_k^2$ is a scaled chi-squared variate with $d$ degrees of freedom. Define $V\triangleq \sqrt{Y}$, then for all $\delta >0$ we have
\begin{align*}
\mathbb{P}\left[ V \ge \sigma \sqrt{d} + \delta \right] \le \exp\left( -\frac{\delta^2}{2\sigma^2}\right).
\end{align*}
\end{lemma}
\begin{Proof} Note that by definition $V(Z_1,\ldots,Z_d)$ is a
$\sigma$-Lipschitz function of $d$ standard normal variables. By Jensen's inequality we have,
\begin{align*}
\mathbb{E}\left[V \right] & \le \sqrt{\mathbb{E}\left[V^2 \right]} = \sigma \sqrt{d}.
\end{align*}
Thus by applying Theorem \ref{t:lipsch} to $V$ we have the claimed bound.
\end{Proof}
%\begin{lemma}[Lemma 9 of \citet{ashwin}] \label{chicon}Let $Z_d$ denote a random variable which is the sum of the squares of $d$ independent $\mathcal{N}(0,\sigma^2)$ Gaussian variables. Then for all $p \in [0,\sigma^2 d]$, we have
%\begin{align*}
%\mathbb{P}\{Z_d \le p \} \le \exp \left(-\frac{d}{2}\left[\log\left(\frac{\sigma^2 d}{p} \right)+\frac{p}{\sigma^2 d} -\frac{1}{p}\right] \right).
%\end{align*}
%\end{lemma}
%\begin{Proof} This is a consequence of the Chernoff bound. In particular, we have for all $\lambda >0$ that
%\begin{align}
%\nonumber \mathbb{P}\{Z_d \le p \} & = \mathbb{P}\{ \exp(-\lambda Z_d) \ge \exp(-\lambda p)\}\\
%\nonumber & \le \exp(\lambda p ) \mathbb{E} [\exp(-\lambda Z_d)]\\
%& = \exp(\lambda p )(1+2\sigma^2\lambda)^{-\frac{d}{2}}, \label{cherno}
%\end{align}
%where in the final step we have used $\mathbb{E}[\exp(-\lambda Z_d)] = (1+2\sigma^2 \lambda)^{-\frac{d}{2}}$, which is valid for all $\lambda > -1/2$. Minimizing the expression for $\lambda >0$ yields the choice $\lambda^* = \frac{1}{2}\left(\frac{d}{p} - \frac{1}{\sigma^2}\right)$ which is greater than $0$ for all $0 < p < \sigma^2 d$. Substituting this choice back into \eqref{cherno} yields the claim.
%\end{Proof}

%\section{Analysis of $\{\ell_1,\ell_2,\ell_{\infty} \}$- MU Selector}
\section{Analysis of Robust Sparse Estimator}
\label{estimator}
Analysis of the $\{\ell_1,\ell_2,\ell_{\infty}\}$-MU Selector
\eqref{belloni} is presented in \citep{bell}, which we adapt here to present guarantees for deterministic (worst case) perturbations to the dictionary.
%\begin{align*}
%\min_{\theta,t,u} \Bigg\{ \lv \theta \rv_1+ \lambda t + \nu u \ \Bigg \lvert \  
%\theta \in \Theta, \Big\lv \frac{1}{d}A^{\top}\big(y-A\theta \big)\Big\rv_{\infty} \le \gamma t + R^2 u, \lv \theta\rv_{2} \le t, \lv \theta \rv_{\infty}\le u \Bigg\}.
%\end{align*}
The analysis in \citep{bell} is in a setting where the error in the
$A$ is random with zero mean. Here, we consider the error to
be deterministic (worst case). Let us start by introducing some
notation and important definitions.

\subsection{Notation and Definitions}
Let $J \subset \{1,\ldots,r \}$ be a set of integers.
%We denote by $\lvert J \rvert$ the carnality of $J$.
For a vector $\theta = (\theta_1,\ldots,\theta_{r}) \in \mathbb{R}^{r}$ we denote by $\theta_{J}$ the vector in $\mathbb{R}^{r}$ whose $j^{th}$ component satisfies $(\theta_{J})_{j} = \theta_{j}$ if $j\in J$, and $(\theta_{J})_j = 0$ otherwise. Let $diag(\cdot)$ be the matrix formed by just the diagonal entries and zeroing out the off diagonal terms. Also let $\Delta \triangleq \hat{\theta} - \theta^{*}$ and $W \triangleq A - A^{*}$, where $\theta^*$ is the true parameter and $A^*$ is the true dictionary without error. 
Define the cone,
\begin{align*}
\mathcal{C}_{J}(u) \triangleq \{ \Delta \in \mathbb{R}^{r} : \lv \Delta_{J^c} \rv_1 \le u \lv \Delta_{J} \rv_1\},
\end{align*}
where $J$ is a subset of $\{1,\ldots, r\}$. For $q \in [1,\infty]$ and an integer $s \in [1,r]$, the $\ell_{q}$-sensitivity (see for example \citet{gautier,MUimprov,bell,belloniconic}) is defined as
\begin{align*}
\kappa_{q}(s,u) \triangleq \min_{J: \lvert J \rvert \le s} \Big(\min_{\Delta \in \mathcal{C}_{J}(u): \lv \Delta \rv_{q} = 1}\frac{1}{d} \lv  A^{*\top}A^* \Delta \rv_{\infty} \Big).
\end{align*}
The $\ell_{q}$-sensitivity is routinely used to study convergence of
estimators under sparsity constraints. If we have $\kappa_{q}(s,u) \ge
c s^{-1/q}$ for some constant $c>0$, this leads to optimal bounds for
the errors. It has also been shown to be a strict generalization of
the restricted eigenvalue property and of the mutual incoherence
condition. Relations between these conditions are provided by Lemma~6
of~\citet{belloniconic}. We restate that lemma here.
\begin{lemma}[Restated from \citet{belloniconic}] \label{1infinity}
Let $u>0$. For any $\alpha \in (0,1)$ there exists a $c>0$ such that for $1\le s \le r$ and $1\le d \le r$ with $\mu/\sqrt{d} \le 1/(cs)$ then
%Let $u>0, 1\le s\le r$. For any $\alpha \in (0,1)$, there exists $c>0$ such that if $\mu \le \sqrt{d}/(cs)$, then
\begin{align*}
\kappa_{\infty}(s,u) \ge \alpha.
\end{align*}
Furthermore, for any $1\le q \le \infty$,
\begin{align*}
\kappa_{q}(s,u) \ge \left(\frac{1}{2s}\right)^{1/q}\kappa_{\infty}(s,u).
\end{align*}
\end{lemma}

Next we highlight the assumptions under which we can establish guarantees for this estimator. 

\subsection{Assumptions}\label{assbell}
We make the following assumptions in the analysis of $\{\ell_1,\ell_2,\ell_{\infty}\}$-MU Selector.
\begin{enumerate}[({D}1)]
\item We assume that the true dictionary $A^*$ is deterministic. We also assume that $A$ is deterministic.
\item We assume that the columns of $A^*$ are normalized, that is, $\lv A^*_{i} \rv_2 = 1 \ \ \forall i\in \{1,2, \ldots, r \}$. 
\item For the matrix $A^*$ we assume the $\ell_{\infty}$-sensitivity is bounded below 
\begin{align*}
\kappa_{\infty}(s,1+\lambda+\nu) \ge 1/4.
\end{align*}
\item We demand that $\lv W \rv_{\infty} \le R$.
\item Finally, the tuning parameters $\lambda$ and $\nu$ are chosen such that
\begin{align*}
8s \underbrace{\left(\frac{\left(\sqrt{s}R^2 + \sqrt{\frac{s}{d}}R\right)\left( 1+\nu + \frac{2\lambda}{\sqrt{s}}\right)}{\lambda}+\frac{R^2(1+\lambda)}{\nu}\right)}_{\triangleq \zeta} \le \frac{1}{2}.
\end{align*}
\end{enumerate}
\begin{remark}\label{rem:kinfinity} If the dictionary $A^*$ is
$\mu/\sqrt{d}$-incoherent and if the sparsity level $s \le C\sqrt{d}/\mu$ for
an appropriate global constant $C$ then by Lemma~\ref{1infinity}
Assumption~(D3) holds for $A^*$.
\end{remark}
\begin{theorem}[Adapted from \cite{bell}] \label{bell}
Let assumptions (D1) - (D5) hold. Assume that the true parameter $\theta^{*}$ is $s-$sparse and belongs to $\Theta$. Let $0 < \lambda,\nu < \infty$, $\gamma = \sqrt{s}R^2 + \sqrt{\frac{s}{d}}R$, and let $\hat{\theta}$ be the $\{\ell_1,\ell_2,\ell_{\infty}\}$-MU Selector. Then
\begin{align*}
\lv \hat{\theta}- \theta^* \rv_{\infty} \le 16(\gamma
 \lv \theta^* \rv_{2}+R^2\lv \theta^* \rv_{\infty}).
\end{align*}
\end{theorem}

\begin{Proof}
Throughout the proof, $J = \{j : \theta^*_{j} \neq 0 \}$.
We proceed in three steps. Step 1 establishes initial
relations and the fact that $\Delta = \hat{\theta} - \theta^*$ belongs
to $\mathcal{C}_{J}(1+\lambda+\nu)$. Step 2 provides a bound on $\frac{1}{d}\lv
A^{\top} A \Delta\rv_{\infty}$. Finally, Step 3 establishes the rate
of convergence stated in the theorem.
We also often use the inequality $\lv \theta
\rv_{\infty} \le \lv \theta \rv_2 \le \lv \theta \rv_1, \forall \theta
\in \mathbb{R}^{r}$.

\emph{Step 1}: We first note that,
\begin{align*}
\frac{1}{d}\left\lv A^{\top}(y - A \theta^*) \right\rv_{\infty} & = \frac{1}{d} \lv A^{\top}W \theta^* \rv_{\infty} \\
&\overset{(i)}{\le} \frac{1}{d}\left\lv A^{* \top} W \theta^* \right\rv_{\infty} + \frac{1}{d}\left\lv W^{\top} W \theta^* \right\rv_{\infty} \\
\small &\overset{(ii)}{\le} \underbrace{\frac{1}{d}\left\lv A^{* \top} W \theta^* \right\rv_{\infty}}_{\triangleq n_1} + \underbrace{\frac{1}{d}\left\lv (W^{ \top} W - diag(W^{\top}W))\theta^* \right\rv_{\infty}}_{\triangleq n_2} \\ & \qquad  \qquad \qquad \qquad  \qquad \qquad + \underbrace{\frac{1}{d}\left\lv diag(W^{\top} W) \theta^* \right\rv_{\infty}}_{\triangleq n_3},
\end{align*}
where both $(i),(ii)$ follow by applications of the triangle inequality. Next we bound $n_1$
\begin{align*}
n_1 & = \frac{1}{d} \lv A^{* \top} W \theta^* \rv_{\infty}.
\end{align*}
Note that the columns of $A^*$ are normalized, $\lv A^*_i \rv_2 =1$ and we have $\lv W \rv_{\infty} \le R$, thus we have all elements of $A^{*\top}W$ are bounded by $\sqrt{d}R$. We also know that $\theta^*$ is $s$-sparse, combining these we get,
\begin{align*}
n_1 & = \frac{1}{d} \lv A^{* \top} W \theta^* \rv_{\infty} \\
& \le \frac{1}{d}(\lv \theta^* \rv_2)(\sqrt{s}\lv A^{*\top}W \rv_{\infty} ) \\
& \le \frac{1}{d}(\lv \theta^* \rv_2)(\sqrt{sd}R) \\
& \le \lv \theta^{*} \rv_2 \left(\sqrt{\frac{s}{d}}R\right),
\end{align*}
where the last step is by Cauchy-Schwartz. Next for $n_2$
\begin{align*}
n_2  & = \frac{1}{d}\left \lv (W^{ \top} W - diag(W^{\top}W))\theta^* \right\rv_{\infty}.
\end{align*}
We know that $\lv W \rv_{\infty} \le R$, thus we have $\lv W^{\top} W - diag(W^{\top}W) \rv_{\infty} \le d R^2$. Again using the fact that $\theta^*$ is $s$-sparse we have,
\begin{align*}
n_2 & = \frac{1}{d}\left \lv (W^{ \top} W - diag(W^{\top}W))\theta^* \right\rv_{\infty} \\
& \le \frac{1}{d}(\lv \theta^* \rv_{2})(\sqrt{s} \lv W^{ \top} W - diag(W^{\top}W) \rv_{\infty})\\
& \le \frac{1}{d}(\lv \theta^* \rv_{2})(\sqrt{s}dR^2)\\
& = \lv \theta^* \rv_2 \sqrt{s}R^2,
\end{align*}
where the first inequality follows by an application of Cauchy-Schwartz. Finally for $n_3$, we again have $\lv W^{\top}W \rv_{\infty} \le dR^2$, thus by Cauchy-Schwartz inequality
\begin{align*}
n_3 = \frac{1}{d}\left\lv diag(W^{\top} W) \theta^* \right\rv_{\infty} \le \lv \theta^* \rv_{\infty} R^2.
\end{align*}
Combining these together we get,
\begin{align}
\label{pshit} \frac{1}{d} \lv A^{\top}(y - A \theta^*) \rv_{\infty} &
\le \left( \sqrt{s}R^2 + \sqrt{\frac{s}{d}}R\right)\lv \theta^*\rv_2 +
R^2  \lv \theta^* \rv_{\infty}.
\end{align}

As $\gamma = \sqrt{s}R^2 + \sqrt{\frac{s}{d}}R$, this implies that $(\theta,t,u) = (\theta^*, \lv \theta^* \rv_2, \lv \theta^* \rv_{\infty})$ is feasible. Let $(\hat{\theta},\hat{t},\hat{u})$ be the optimal solution, then we have
\begin{align*}
\lv \hat{\theta} \rv_1 + \lambda  \lv \hat{\theta} \rv_2 + \nu  \lv \hat{\theta} \rv_{\infty} \le \lv \hat{\theta} \rv_1   +  \lambda  \hat{t} + \nu \hat{u} \le  \lv \theta^* \rv_1 + \lambda \lv \theta^* \rv_2 + \nu \lv \theta^* \rv_{\infty}.
\end{align*}
By rearranging terms and by triangle inequality we get the relation
\begin{align*}
\lv \hat{\theta}_{J^C} \rv_1 \le (1+ \lambda +\nu ) \lv \hat{\theta}_{J} - \theta^* \rv_1 = (1+ \lambda +\nu ) \lv \Delta_{J} \rv_1.
\end{align*}
This proves that $\Delta \in \mathcal{C}_{J}(1+\lambda+\nu)$. Also by similar arguments we get
\begin{align}
\label{tshit} \hat{t} - \lv \theta^* \rv_2 \le \frac{\lv \Delta \rv_1 + \nu \lv \Delta \rv_{\infty}}{\lambda} \le \frac{(1+\nu)\lv \Delta \rv_1}{\lambda} \\
\label{ushit} \text{and, } \hat{u} - \lv \theta^* \rv_{\infty} \le \frac{\lv \Delta \rv_1+ \lambda \lv \Delta \rv_2}{\nu} \le \frac{(1+ \lambda)}{\nu} \lv \Delta \rv_1 .
\end{align}

\emph{Step 2}: By applications of the triangle inequality we have
\begin{align*}
\frac{1}{d} \lv A^{* \top} A^* \Delta \rv_{\infty} &\le \frac{1}{d} \left[\lv A^{\top}A^* \Delta \rv_{\infty} + \lv W^{\top}A^* \Delta\rv_{\infty} \right]\\
& \le \frac{1}{d} \left[ \lv A^{\top}A \Delta \rv_{\infty} + \lv A^{\top}W \Delta \rv_{\infty} + \lv W^{\top}A^* \Delta \rv_{\infty} \right]\\
& \le \frac{1}{d} \left[ \underbrace{\lv A^{\top}(y-A\theta^*)\rv_{\infty}}_{\triangleq m_1}+ \underbrace{\lv A^{\top}(y-A\hat{\theta})\rv_{\infty}}_{\triangleq m_2}+ \underbrace{\lv A^{\top}W \Delta \rv_{\infty}}_{\triangleq m_3} + \underbrace{\lv W^{\top}A^*\Delta \rv_{\infty}}_{\triangleq m_4}\right].
\end{align*}

Now we bound each of these terms
\begin{align*}
m_1 & \overset{(i)}{\le}  d(\gamma \lv \theta^* \rv_2 + R^2 \lv \theta \rv_{\infty})\\
m_2 &\overset{(ii)}{\le}  d(\gamma \hat{t} + R^2 \hat{u}) \le d \left( \gamma \lv \theta^* \rv_2 + R^2 \lv \theta^* \rv_{\infty}+ \left\{ \frac{\gamma(1+\nu)}{\lambda} +\frac{R^2(1+ \lambda)}{\nu}\right\} \lv \Delta \rv_1 \right)\\
m_3 & \overset{(iii)}{\le}   \left(dR^2 + \sqrt{d}R\right)\lv \Delta \rv_{1}+d R^2  \lv \Delta \rv_{\infty}\\
m_4 & \overset{(iv)}{\le}   \sqrt{d}R\lv \Delta \rv_1,
\end{align*}
where $(i)$ follows as $(\theta^*,\lv \theta^* \rv_2,\lv \theta^*\rv_{\infty})$ is a feasible point, $(ii)$ is because $(\hat{\theta},\hat{t},\hat{u})$ is a (optimal) feasible point along with \eqref{tshit}, \eqref{ushit}. Bound $(iii)$ follows by similar arguments made to arrive at \eqref{pshit} and finally $(iv)$ is due to H\"older's inequality.  Combing these we have the following bound 
\begin{align*}
\frac{1}{d} \lv A^{* \top} A^* \Delta \rv_{\infty} & \le 2 \gamma \lv \theta^* \rv_{2} + 2R^2 \lv \theta^* \rv_{\infty} + \left\{ \frac{\gamma(1+\nu)}{\lambda} + \frac{R^2(1+\lambda)}{\nu}\right\}\lv \Delta \rv_1 \\ & \qquad \qquad \qquad + \frac{R}{\sqrt{d}} \lv \Delta \rv_1 + R^2 \lv \Delta \rv_{\infty} + \left( R^2 + \frac{R}{\sqrt{d}}\right)\lv \Delta \rv_1.
\end{align*}
Simplifying using $\lv \Delta \rv_{\infty} \le \lv\Delta \rv_1$ and $\gamma =\sqrt{s}\left(R^2 + \frac{R}{\sqrt{d}}\right)$ we get
\begin{align}\label{mainb}
\frac{1}{d} \lv A^{* \top} A^* \Delta \rv_{\infty} & \le 2 \gamma \lv \theta^* \rv_{2} + 2R^2 \lv \theta^* \rv_{\infty} + \left( \gamma \frac{\left(1+ \nu +\frac{2\lambda}{\sqrt{s}}\right)}{\lambda} + \frac{R^2 (1+ \lambda)}{\nu}\right)\lv\Delta \rv_1.
\end{align}
\emph{Step 3}: Define 
\begin{align*}
\zeta \triangleq \left( \gamma\frac{\left(1+ \nu + \frac{2\lambda}{\sqrt{s}} \right)}{\lambda}+ \frac{R^2(1+\lambda)}{\nu}\right).
\end{align*}
Rewriting \eqref{mainb} using the definition of $\zeta$ we have
\begin{align*}
\frac{1}{d}\lv A^{*\top}A^* \Delta \rv_{\infty} \le 2 \gamma \lv \theta^* \rv_{2}+ 2R^2\lv \theta^* \rv_{\infty} + \zeta  \lv \Delta \rv_1.
\end{align*}
By the assumption on $\ell_{\infty}$-sensitivity and Lemma \ref{1infinity} we have $\kappa_1(s,u) \ge \frac{1}{2s} \kappa_{\infty}(s,u) \ge \frac{1}{8s}$. %assumed that \kappa_{\infty}(s,u) \ge 1/4.
Thus by definition of $\ell_1$-sensitivity we have
\begin{align*}
\frac{1}{d}\lv A^{*\top} A^* \Delta\rv_{\infty} \ge \kappa_1(s,1+\lambda+\nu)\lv \Delta \rv_1.
\end{align*} 
Combining this with the previous display gives us
\begin{align*}
\frac{1}{d}\lv A^{* \top}A^* \Delta\rv_{\infty} & \le 2 \gamma \lv \theta^* \rv_{2}+ 2R^2\lv \theta^* \rv_{\infty} + \frac{\zeta}{\kappa_{1}(s,1+\lambda+\nu)}\left(\frac{1}{d}\lv A^{*\top}A^* \Delta\rv_{\infty}\right)\\
& \le 2 \gamma \lv \theta^* \rv_{2}+ 2R^2\lv \theta^* \rv_{\infty} + 8 s \zeta\left(\frac{1}{d}\lv A^{*\top}A^* \Delta\rv_{\infty}\right).
\end{align*}
By assumption (D5) -- $8s \zeta \le 1/2$, therefore we have the claimed error bound
\begin{align*}
\frac{1}{2d} \lv A^{* \top}A^* \Delta\rv_{\infty} &\le 2 \gamma \lv \theta^* \rv_{2}+ 2R^2\lv \theta^* \rv_{\infty} \\
\kappa_{\infty}(s,1+\lambda+\nu) \lv \Delta \rv_{\infty} & \le 4 \gamma \lv \theta^* \rv_{2}+ 4R^2\lv \theta^* \rv_{\infty} \\
\lv \hat{\theta} - \theta^* \rv_{\infty} & \le 16(\gamma \lv \theta^* \rv_{2}+R^2\lv \theta^* \rv_{\infty}).
\end{align*}
\end{Proof}

\section{Lower Bounds: Proof of Theorem \ref{minimaxtheo}}
\label{minimaxsec}In this section we will show that when the uncertainty in the dictionary measured in matrix infinity norm scales as $R = \mathcal{O}(1/\sqrt{s})$, the $\{\ell_1, \ell_2, \ell_{\infty} \}$-MU Selector is information theoretically optimal up to logarithmic factors and the infinity norm of the error (in the worst case) is lower bounded by $C R \lv \theta^* \rv_{2}$. We will prove this by Fano's method (see for example review in \citet{fano, tsybakov2009introduction}). The proof technique to show this estimator is minimax optimal is adapted from \citet{belloniconic}. We define the sets
\begin{align*}
B_{0}(s) = \{\theta : \lv \theta \rv_{0} \le s \} \qquad \text{and } \qquad S_{2}(L) = \{\theta : \lv \theta \rv_{2} = L \},
\end{align*}
where $L > 0$. We define the parameter set to be $\Theta = B_0(s) \cap S_2(L)$, which is the set of $s-$sparse vectors with $\lv\cdot \rv_2$ norm equal to $L$. To prove this theorem we will choose a particular probability distribution over the set of underlying \emph{true dictionaries} $\mathbb{P}_{A^*}$ and also a distribution over the deviations from the true dictionary $\mathbb{P}_{W}$. We will assume that the entries of $A^*$ are drawn i.i.d. from a zero-mean Gaussian distribution $\mathcal{N}(0,\sigma_D^2)$ and the entries of $W$ are chosen i.i.d. from a zero mean Gaussian distribution $\mathcal{N}(0,\sigma_E^2)$ independent of the distribution generating $A^*$. We set $\sigma_D = \mathcal{O}(1/\sqrt{d})$ and $\sigma_E = \mathcal{O}(R/\sqrt{log(dr)})$. We now restate a formal version of Theorem \ref{minimaxtheo}. 
\begin{theorem} Let $r\ge 2$, $2\le s \le r$, and $L>0$. Let $y = A^*\theta^*$ where  $A^* \in \mathbb{R}^{d \times r}$ and $\theta^*$ is a $s$-sparse vector with norm $\lv \theta^* \rv_2 = L$. Further let the entries of $A^*$ be drawn from $\mathcal{N}(0,\sigma_D^2)$ and independently let the entries of the perturbation $W$ be drawn from the distribution $\mathcal{N}(0,\sigma^2_E)$. Let $A = A^* + W$, $\sigma_{D}^2  = \mathcal{O}(1/d)$ and $\sigma_E^2 = R/\log(dr)$. Then there exists constants $C$ and $C'>0$ such that
\begin{align*}
\inf_{\hat{T}} \sup_{\theta \in B_0(s) \cap S_2(L)} \mathbb{P}_{A^*,W} \left[\lv \hat{T} - \theta \rv_{\infty} \ge C R L\sqrt{1- \frac{\log(s)}{\log(r)}} \right] > C',
\end{align*}
where $\inf_{\hat{T}}$ denotes the infimum over all measurable estimators $\hat{T}$ with input $(y,A,R)$.
\end{theorem}
%\begin{remark}Note that when $R  = \mathcal{O}(1/\sqrt{s})$, this lower bound matches the upper bound we have for Theorem \ref{bell} (up to logarithmic factors) and hence is minimax optimal.\end{remark}
\begin{Proof} We define a finite set of ``hypotheses" (packing set) included in $B_0(s) \cap S_2(L)$. To this end, we first introduce
\begin{align*}
\mathcal{M} = \{x \in \{0,1 \}^{r-1} : \rho_{H}(\boldsymbol{0},x) = s-1 \},
\end{align*}
where $\rho_H$ denotes the Hamming distance between elements of $\{0,1 \}^{r-1}$, and $\boldsymbol{0}$ is the zero vector. Then there exists a subset $\mathcal{M}'$ of $\mathcal{M}$ such that for any $x,x'$ in $\mathcal{M}'$ with $x\neq x'$, we have $\rho_H(x,x') > s/16$ and moreover the cardinality of $\mathcal{M}'$ is bounded below
\begin{align*}
\log\lvert \mathcal{M}' \rvert \ge C s \log\left(\frac{r}{s}\right),
\end{align*}
for some constant $C$. This follows from Varshamov-Gilbert bound (see Lemma 2.9 in \citet{tsybakov2009introduction}) if $s-1 > (r-1)/2$ and from Lemma A.3 in \citet{rigo} if $s-1 \le (r-1)/2$. We denote $\omega_j'$ to be the elements of the finite set $\mathcal{M}'$. For $j=1,\ldots,\lvert \mathcal{M} \rvert$, we define the vectors $\omega_j \in \{0,1 \}^{r}$ with components $\omega_{j1} = 0$ and $\omega_{jk} = \omega_{j(k-1)}'$ for $k>2$, where $\omega_{jk}$ is the $k$-th component of $\omega_j$. We also define $\omega_0$ as the vector in $\{0,1 \}^r$ with all components equal to $0$ except the first one equal to 1. We now define the set of ``hypotheses" (packing set of $\Theta$) ($\bar{\omega}_j, j = 0,\ldots, \lvert \mathcal{M}'\rvert+1$), where $\bar{\omega}_0 = R \omega_0$ and
\begin{align*}
\bar{\omega}_j = \frac{L}{\sqrt{1+ \psi^2(s-1)}}(\omega_0 + \psi \omega_j), \qquad j = 1,\ldots, \lvert \mathcal{M}' \rvert +1.
\end{align*} 
Here $\psi$ is a positive parameter that will be chosen appropriately. Note that these vectors are $s$-sparse and have $\lv \bar{\omega}_j \rv_2 = L$. By Lemma \ref{KLlem} we have the KL divergence is bounded, 
\begin{align*}
\mathcal{K}(\mathbb{P}_{\bar{\omega}_j},\mathbb{P}_{\bar{\omega}_0}) & = \frac{d \sigma_D^2}{2 \sigma_{E}^2 \lv \bar{\omega}_0 \rv_2^2} \lv \bar{\omega}_j - \bar{\omega}_0 \rv^2\\
& \le \frac{d\sigma_D^2}{2 \sigma_{E}^2 L^2} \left(\frac{\psi^2 L^2 s}{1+ \psi^2 (s-1)} \right)\\
& \le \psi^2\left(\frac{ s d\sigma_D^2}{2 \sigma_{E}^2(1+ \psi^2(s-1)) }\right).
\end{align*}
If we choose $\psi = C \sqrt{\frac{\sigma_E^2 \log(r/s)}{d \sigma_D^2}}$ with $C$ being an appropriately chosen constant independent of dimensions ($s,d,r$) and $L$ we get that for all $j$,
\begin{align*}
\mathcal{K}(\mathbb{P}_{\bar{\omega}_j},\mathbb{P}_{\bar{\omega}_0}) \le \frac{1}{16} \log(\lvert \mathcal{M}' \rvert).
\end{align*} 
Thus for $j$ and $j'$ both different from 0,
\begin{align*}
\lv \bar{\omega}_{j} - \bar{\omega}_{j'} \rv_{\infty} = \frac{L \psi}{\sqrt{1+ \psi^2(s-1)}} \ge C\frac{L \sigma_E \sqrt{\log(r/s)}}{\sqrt{d}\sigma_D},
\end{align*}
and for $j \neq 0 $ we have
\begin{align*}
\lv \bar{\omega}_{j} - \bar{\omega}_{0} \rv_{\infty} \ge \frac{L \psi \lv \omega_j \rv_{\infty}}{\sqrt{1+ \psi^2(s-1)}} \ge  C\frac{L \sigma_E \sqrt{\log(r/s)}}{\sqrt{d}\sigma_D}.
\end{align*}
We want the columns of $\lv A^* \rv_{2} \le 1$ (upper bound used in the proof of Theorem \ref{bell}), hence we want $\sigma_D  = \mathcal{O}(1/\sqrt{d})$ (this follows by an application of Lemma \ref{chicon} followed by a union bound over the $r$ columns using the fact that $r = \mathcal{O}(poly(d))$). We also demand that our deviation from the true dictionary be bounded by $R$ with high probability over all entries so we choose $\sigma_E \le \mathcal{O}( R/\sqrt{log(dr)})$. Hence given our choices of $\sigma_E$ and $\sigma_{D}$ we have for any $j,j'$
\begin{align*}
\lv \bar{\omega}_{j} - \bar{\omega}_{j'} \rv_{\infty} \ge C L R \left( \sqrt{1-  \frac{\log(s)}{\log(r)}}\right).
\end{align*}
We can now apply Theorem 2.7 in \citet{tsybakov2009introduction} to complete the proof. 
\end{Proof}

\begin{lemma}\label{KLlem}Let $\theta_1 \in \mathbb{R}^r$ and $\theta_2 \in \mathbb{R}^r$ be such that $\lv \theta_1 \rv_2 = \lv \theta_2 \rv_2$. Under the assumptions stated in the Appendix \ref{minimaxsec} we have
\begin{align*}
\mathcal{K}(\mathbb{P}_{\theta_1},\mathbb{P}_{\theta_2}) = \frac{d\sigma_D^2}{2 \sigma_{E}^2 \lv \theta_2 \rv_2^2} \lv \theta_1 - \theta_2 \rv^2.
\end{align*}
\end{lemma}
\begin{Proof} By the properties of Kullback Leibler divergence between product measures, it suffices to prove the lemma for $d=1$. Let $\theta \in \mathbb{R}^r$. Consider the random vector $(U,V)$ where
\begin{align*}
V = (D_1 + E_1,\ldots, D_r +E_r),
\end{align*} 
with $D = (D_1,D_2,\ldots,D_r)^{\top}$ a zero-mean Gaussian vector with covariance $\sigma^2_D I_{r\times r}$ and $E = (E_1,E_2,\ldots,E_r)^{\top}$ a zero mean Gaussian vector with covariance $\sigma^2_E I_{r\times r}$ independent of A and 
\begin{align*}
U = \sum_{j = 1}^{r} \theta_j (V_j - E_j).
\end{align*}
We introduce some variables 
\begin{align*}
\tilde{\Sigma} = \frac{\sigma^2_E}{\sigma^2_D + \sigma^2_E} I_{r \times r}, \quad \Pi = \frac{\sigma_D^2}{\sigma_D^2 + \sigma_E^2} I_{r \times r}, \quad c_{\theta} = \theta^{\top} \Pi \theta = \frac{\sigma_D^2}{\sigma_D^2 + \sigma_E^2} \lv \theta \rv_2^2.
\end{align*}
We find the conditional distribution $\mathcal{L}_{\theta}(U \lvert V)$ of $U$ given $V$. Also note that the vector $(V_1,\ldots,V_r,E_1,\ldots,E_r)^{\top}$ is a zero-mean Gaussian random vector with covariance matrix
\begin{align*}
\begin{pmatrix}
  (\sigma^2_D + \sigma^2_E)I_{r\times r} & \sigma^2_E I_{r \times r}  \\
  \sigma^2_E I_{r \times r} & \sigma^2_E I_{r \times r}
 \end{pmatrix}.
\end{align*}
So that $\mathcal{L}_{\theta}(E \lvert V)$ is a Gaussian with mean $\tilde{\Sigma} V$ and covariance $\sigma_E^2(I_{r\times r} - \tilde{\Sigma})$. This implies that $\mathcal{L}_{\theta}(U\lvert V)$ is Gaussian with mean $\theta^{\top} \Pi \theta$ and variance $c_{\theta} \sigma_E^2$. Then the logarithm of density of $\mathcal{L}_{\theta}(U\lvert V)$, denoted by $\ell_{\theta}(U\lvert V)$ satisfies 
\begin{align*}
\ell_{\theta} (U \lvert V) = -\frac{1}{2} \log(2\pi) - \frac{1}{2}\log(c_{\theta}\sigma_E^2) - \frac{1}{2c_{\theta}\sigma_E^2} (U - \theta^{\top}\Pi V)^2.
\end{align*} 
Now let $\theta_1 \in \mathbb{R}^r$ and $\theta_2 \in \mathbb{R}^r$ with $\lv \theta_1 \rv_2 = \lv \theta_2 \rv_2$. Then,
\begin{align*}
\ell_{\theta_1}(U \lvert V) - \ell_{\theta_2}(U \lvert V) & = \frac{1}{2} \underbrace{\left( \log \left(\frac{c_{\theta_2}}{c_{\theta_1}}\right)\right)}_{=0} + \frac{1}{2c_{\theta_2}\sigma_E^2}\left( (U - \theta_2^{\top}\Pi V)^2 - (U - \theta_1^{\top} \Pi V)^2\right) \\& \qquad \qquad \qquad \qquad + \underbrace{\left(\frac{1}{2c_{\theta_2}\sigma_E^2} - \frac{1}{2c_{\theta_1}\sigma_E^2}\right)}_{=0} (U - \theta_1^{\top} \Pi V)^2\\
& = \frac{1}{2c_{\theta_2}\sigma_E^2}\left( (U - \theta_2^{\top}\Pi V)^2 - (U - \theta_1^{\top} \Pi V)^2\right).
\end{align*}
Since the distribution of $V$ does not depend on $\theta_1$ we obtain that in the case $d=1$,
\begin{align*}
\mathcal{K}(\mathbb{P}_{\theta_1},\mathbb{P}_{\theta_2}) & = \frac{1}{2c_{\theta_2}\sigma_E^2} \mathbb{E}_{\theta_1}\left[(U - \theta_2^{\top}\Pi V)^2 - (U - \theta_1^{\top}\Pi V)^2  \right]\\
& = \frac{\sigma_D^2 + \sigma_E^2}{2\sigma_E^2 \sigma_D^2 \lv \theta \rv_2^2} \Big[\sigma_D^2(\theta_1^{\top} - \theta_2^{\top}\Pi)I_{r \times r}(\theta_1 - \Pi \theta_2) \\ & \qquad \qquad - \sigma_{D}^2(\theta_1^{\top} - \theta_1^{\top}\Pi)I_{r \times r}(\theta_1 - \Pi \theta_1) + (\theta_2^{\top} \Pi^2 \theta_2 - \theta_1^{\top} \Pi^2 \theta_1)\Big].
\end{align*}
Where in the final step the cross terms are zero by the independence of $D$ and $E$. Developing this expression leaves us with
\begin{align*}
\mathcal{K}(\mathbb{P}_{\theta_1},\mathbb{P}_{\theta_2}) & = \frac{\sigma_D^2 + \sigma_E^2}{2 \sigma_E^2 \sigma_D^2 \lv \theta \rv_2^2}\left[ (\theta_1 - \theta_2)^{\top} \Pi \sigma_{D}^2 I_{r \times r} (\theta_1 - \theta_2) \right]\\
& = \frac{\sigma_D^2}{2 \sigma_{E}^2 \lv \theta_2 \rv_2^2} \lv \theta_1 - \theta_2 \rv^2.
\end{align*}
\end{Proof}
\end{document}